\documentclass{article} 
\usepackage{iclr2023_conference,times}
\usepackage{comment}

\usepackage{graphicx}
\usepackage{comment}
\usepackage{times}
\usepackage{epsfig}
\usepackage{amsmath}
\usepackage{amssymb}
\usepackage{url}
\usepackage{multirow}
\usepackage{subfigure}
\usepackage{color}
\usepackage{hyperref}
\usepackage{blindtext} 
\usepackage{epigraph} 
\usepackage{xspace}

\usepackage{multirow}
\usepackage{amssymb}
\usepackage{pifont}

\usepackage{nicefrac}       
\usepackage{microtype}      

\usepackage{wrapfig}
\usepackage{enumitem}
\usepackage{booktabs}
\usepackage{floatrow}
\usepackage[ruled]{algorithm2e}
\usepackage{listings}
\usepackage{enumitem}
\usepackage{wrapfig}
\usepackage{lipsum}
\usepackage{graphics}

\newfloatcommand{capbtabbox}{table}[][\FBwidth]

\hypersetup{
    colorlinks=true,
    linkcolor=blue,
    filecolor=magenta,      
    urlcolor=cyan,
}
\makeatletter
\newcommand*{\etc}{%
    \@ifnextchar{.}%
        {\textit{etc}}%
        {\textit{etc.}\@\xspace}%
}
\makeatother

\usepackage{amsmath,amsfonts,bm}

\def\eqref#1{equation~\ref{#1}}

\def\1{\bm{1}}

\DeclareMathAlphabet{\mathsfit}{\encodingdefault}{\sfdefault}{m}{sl}
\SetMathAlphabet{\mathsfit}{bold}{\encodingdefault}{\sfdefault}{bx}{n}

\newcommand{\model}{\text{Analogical Networks}}
\newcommand{\modelnospace}{\text{Analogical Networks}}
\renewcommand{\cite}{\citep}

\newcommand{\partnetbl}{\texttt{PartNet}}

\newcommand{\modelgtret}{\texttt{AnalogicalNets} \texttt{OrclCategoryRetrv}}

\newcommand{\singleabl}{\texttt{AnalogicalNets single-mem}}
\newcommand{\multiabl}{\texttt{AnalogicalNets multi-mem}}
\newcommand{\nopretrain}{\texttt{AnalogicalNets} \texttt{w/o within-scene}}
\newcommand{\detrbl}{\texttt{DETR3D}}
\newcommand{\detrblnospace}{\texttt{DETR3D}}

\newcommand{\protobl}{\texttt{PrototypicalNets}}

\newcommand{\memtransfabl}{\texttt{Re-DETR3D}}

\newcommand{\nopretrainsingle}{\singleabl{} \texttt{w/o within-scene}}

\newcommand{\printfnsymbol}[1]{%
  \textsuperscript{\@fnsymbol{#1}}%
}

\title{Analogy-Forming Transformers for Few-Shot 3D Parsing}

\iclrfinalcopy
\author{Nikolaos Gkanatsios\thanks{Equal contribution} 
 , Mayank Singh$^*$, Zhaoyuan Fang,\\ \textbf{Shubham Tulsiani} \&
\textbf{Katerina Fragkiadaki}
\\
School of Computer Science\\
Carnegie Mellon University \\
Pittsburgh, PA 15213, USA \\
\texttt{\{ngkanats,mayanks2,zhaoyuaf,stulsian,katef\}@cs.cmu.edu} \\
}

\newcommand{\katef}[1]{{\color{magenta} }}

\begin{document}

\maketitle

\begin{abstract}
We present \model{}, a model that encodes domain knowledge explicitly, in  a  collection of structured labelled 3D scenes, in addition to  implicitly,  as model   parameters, and segments 3D object scenes with analogical reasoning: instead of mapping a scene to part segments directly, 
our model first retrieves related scenes from memory and their corresponding part  structures, and then predicts \textit{analogous} part structures  for the input scene, via an end-to-end learnable modulation mechanism. By conditioning on more than one retrieved memories, compositions of  structures are predicted, that mix and match parts across the retrieved memories.   One-shot, few-shot or many-shot learning are treated  uniformly in \model{}, by conditioning on the appropriate set of memories, whether taken from a single, few or many memory exemplars, and inferring analogous parses. We show \model{} are competitive with state-of-the-art 3D segmentation  transformers in many-shot settings, and outperform them, as well as existing paradigms of meta-learning and few-shot learning, in few-shot settings. \model{} successfully segment instances of novel object categories simply by expanding their memory, without any weight updates. Our code and models are publicly available in the project webpage: \url{http://analogicalnets.github.io/}.
\end{abstract}
\epigraph{Ask not what it is, ask what it is like.}{\textit{Moshe Bar}}

\section{Introduction}

The dominant paradigm in existing deep visual learning is to train high-capacity networks that map input observations to task-specific outputs.  
Despite their success across a plethora of tasks, these models struggle to perform well in \emph{few-shot} settings where only a small set of examples are available for learning. Meta-learning approaches provide one promising solution to this by enabling efficient task-specific adaptation of generic models, but this specialization comes at the cost of poor performance on the original tasks as well as the need to adapt separate models for each novel task.

\begin{figure}[h!]
  \centering
    \includegraphics[width=0.95\linewidth]{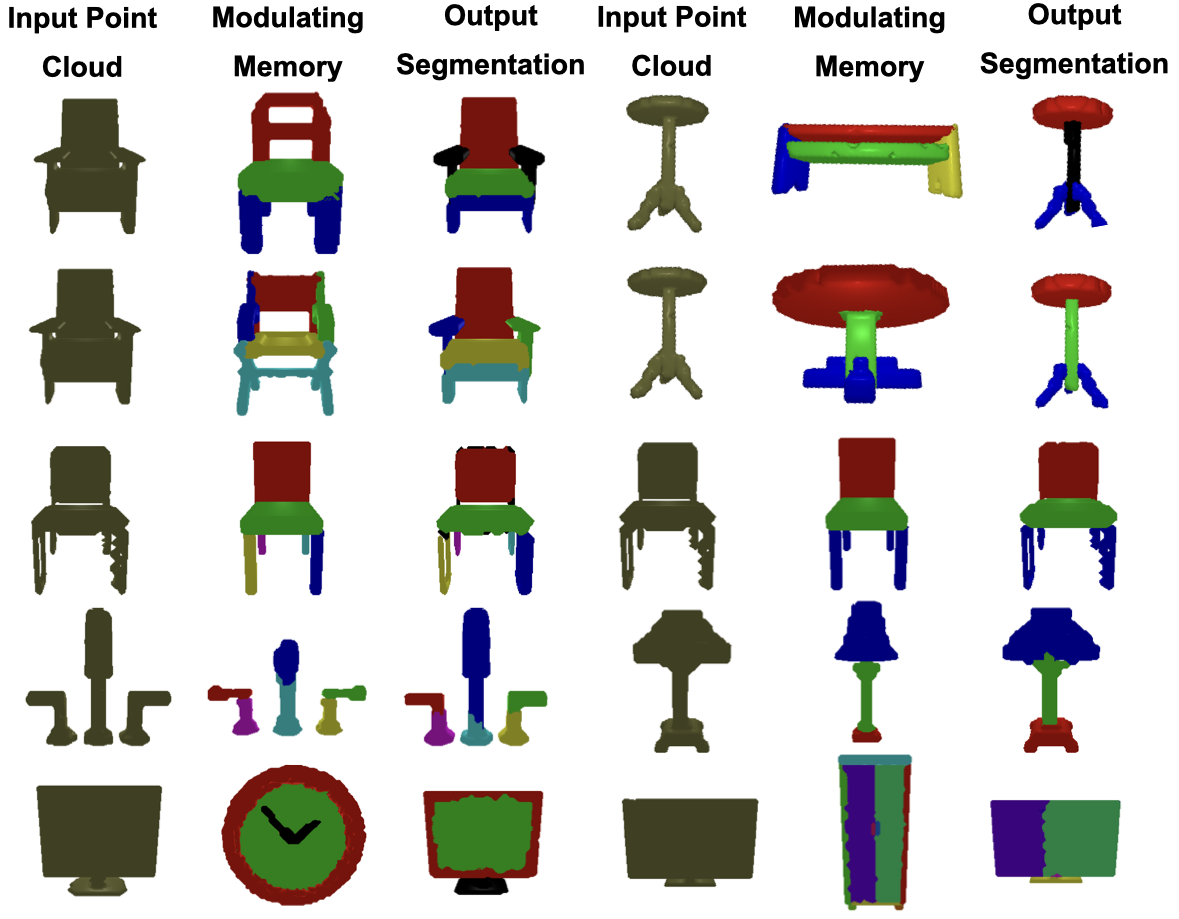}
  \caption{  \textbf{\model{} form analogies between retrieved memories and the input scene} by using memory part encodings  as queries to localize corresponding parts in the  scene. 
  Retrieved  memories (2nd and 5th columns)  modulate segmentation of the  input 3D point cloud (1st and 4th columns, respectively).   We indicate corresponding parts between the memory and the input scene with the same color.  Cross-object part correspondences emerge even without any part association or semantic part labelling supervision (5th row). 
  For example, the model learns to correspond the parts of a clock and a TV set,  without ever trained with such cross scene part correspondence. Parts shown in black in columns 3 and 6 are decoded from scene-agnostic queries and thus they are not in correspondence to any parts of the memory scene. Conditioning the same input point cloud on memories with finer or coarser labellings results in  segmentation of analogous granularity (3rd row).
  }
  \label{fig:intro}
\end{figure}

\katef{what we propose}
We introduce \modelnospace, a semi-parametric learning framework for 3D scene parsing that pursues analogy-driven  prediction: instead of mapping the input scene to part segments directly, the model reasons analogically and maps  the input to modifications and compositions of past labelled visual  experiences. \model{}  encode domain knowledge explicitly in  a  collection of structured  labelled scene memories  as well as implicitly, in model parameters. Given an input 3D scene, the model retrieves relevant memories and uses them to modulate inference and segment object parts in the input point cloud. 
During modulation,  the input scene and the retrieved memories  are jointly encoded and contextualized via cross-attention operations. 
The contextualized memory part features are then used to segment analogous parts in the 3D input scene,  binding the predicted part structure to the one from memory, as shown in Figure~\ref{fig:intro}.   Given the same input scene, the output prediction changes with varying conditioning  memories.  For example, conditioned on visual memories of varying label granularity, the model segments the input in a granularity analogous to the one of the retrieved memory. One-shot, few-shot or many-shot learning are treated  uniformly in \model{}, by conditioning on the appropriate set of memories. 
This is very beneficial over methods that specifically target few-shot only scenarios,  since, at test time, an agent usually cannot know whether a scene is an example of many-shot or few-shot categories. 

\model{} learn to bind memory part features to input scene part segments. Fine-grained part correspondence annotations across two 3D scenes are not easily available. 
We devise a novel within-scene pre-training scheme to encourage correspondence learning across  scenes. We augment (rotate and deform) a given scene in two distinct ways, and train the modulator to parse one of them given the other as its modulating memory, bypassing the retrieval process. During this within-scene training, we have access to the part correspondence between the memory and the input scene, and we use it to supervise the query-to-part assignment process. We show within-scene training helps our model learn to associate  memory queries to similarly labelled parts \textit{across scenes} 
without ever using cross-scene part correspondence annotations, as shown in Figure~\ref{fig:intro}.

We test our model on the PartNet benchmark of \citet{Mo_2019_CVPR} for 3D object segmentation. We compare against state-of-the-art (SOTA) 3D  object segmentors,  as well as  meta-learning and few-shot learning \cite{NIPS2017_cb8da676} baselines adapted for the task of 3D parsing. Our experiments show that \model{}   perform similarly to the parametric alone baselines in the standard many-shot  train-test split and particularly shine over the parametric  baselines in the few-shot setting:  \model{} segment novel instances much better than parametric  existing models,  simply by expanding the memory repository with encodings of a few exemplars,  even without any weight updates. We further compare against variants of our model that consider memory retrieval and  attention without memory query binding, and thus lack explicit analogy formation, as well as other ablative versions of our model to quantify the contribution of the retriever and the proposed within-scene memory-augmented pre-training. Our code and models are publicly available in the project webpage: \url{http://analogicalnets.github.io/}.

\section{Related work}
\paragraph{Few-shot prediction: meta-learning and learning associations} A key goal for our approach is to enable accurate inference in few-shot settings. Previous approaches \cite{maml_finn, rusu2018meta, NIPS2017_cb8da676, KGfewshot, bar2022visual, Mangla_2020_WACV, wang2020frustratingly, nguyen2019feature, liu2020part, tian2020prior, yang2020prototype} that target similar settings can be broadly categorized as relying on either meta-learning or learning better associations. Meta-learning approaches tackle few-shot prediction by learning a generic model that can be efficiently adapted to a new task of interest from a few labelled examples. While broadly applicable, these methods result in catastrophic forgetting of the original task during adaptation and thus require training a new model for each task of interest. Moreover, the goal of learning generic and rapidly adaptable models can lead to suboptimal performance over the base tasks with abundant data. An alternative approach for the few-shot setting is to learn better associations. For example, the category of a new example may be inferred by transferring the label(s) from the one (or few) closest samples \citep{NIPS2017_cb8da676, Sung_relationnet_2018_CVPR}. While this approach obviates the need for adapting models and can allow prediction in few-shot and many-shot settings, the current approaches are only applicable to global prediction, e.g., image labels. Our work can be viewed as extending such association-based methods to allow predicting fine-grained and generic visual structures using our proposed modulation-based prediction mechanism.

\textbf{Memory-augmented neural networks} 
\model{} is a type of memory-augmented neural networks. 
Memory-augmentation of parametric models permits  \textbf{fast learning} with few examples, where  data are saved in and can be accessed from the memory immediately after their acquisition \cite{MEMORYAUGMENTED}, while learning via parameter update is slow and requires multiple gradient iterations on de-correlated examples. 
External memories have recently been used to scale up language models \cite{retro,NNLM}, and alleviate from the limited context window of parametric transformers \cite{https://doi.org/10.48550/arxiv.2203.08913}, as well as to store knowledge in the form of entity mentions \cite{DBLP:journals/corr/abs-2110-06176}, knowledge graphs \cite{https://doi.org/10.48550/arxiv.2202.10610} and question-answer pairs \cite{https://doi.org/10.48550/arxiv.2204.04581}. Memory attention layers in these models are 
used to influence the computation of transformer
layers and have proven very successful for factual question answering, but also for sentence completion over their parametric counterparts.

\paragraph{In-context learning }
In-context learning (ICL)  \cite{DBLP:journals/corr/abs-2005-14165} aims to induce a model to perform a task by feeding in input-output examples  along with an unlabeled query example. 
The primary advantage of in-context learning is that it enables a single model to perform many tasks immediately without fine-tuning. 
\model{} are in-context learners in that they infer the part segmentation of an object 3D point cloud in the context of retrieved  labelled object 3D point clouds. While in language models ICL emerges at test time while training unsupervised (and out-of-context) for language completion, \model{} are trained  in-context with related examples using supervision and self-supervision. Although in prompted language models \cite{https://doi.org/10.48550/arxiv.2207.05608,DBLP:journals/corr/abs-2005-14165} the input-output pairs are currently primarily decided by the engineer, in \model{} they are automatically inferred by the retriever.

\section{ \model{} for 3D object  parsing}  \label{sec:prelim}

\begin{figure}[t!]
  \centering
  \includegraphics[width=1.0\linewidth]{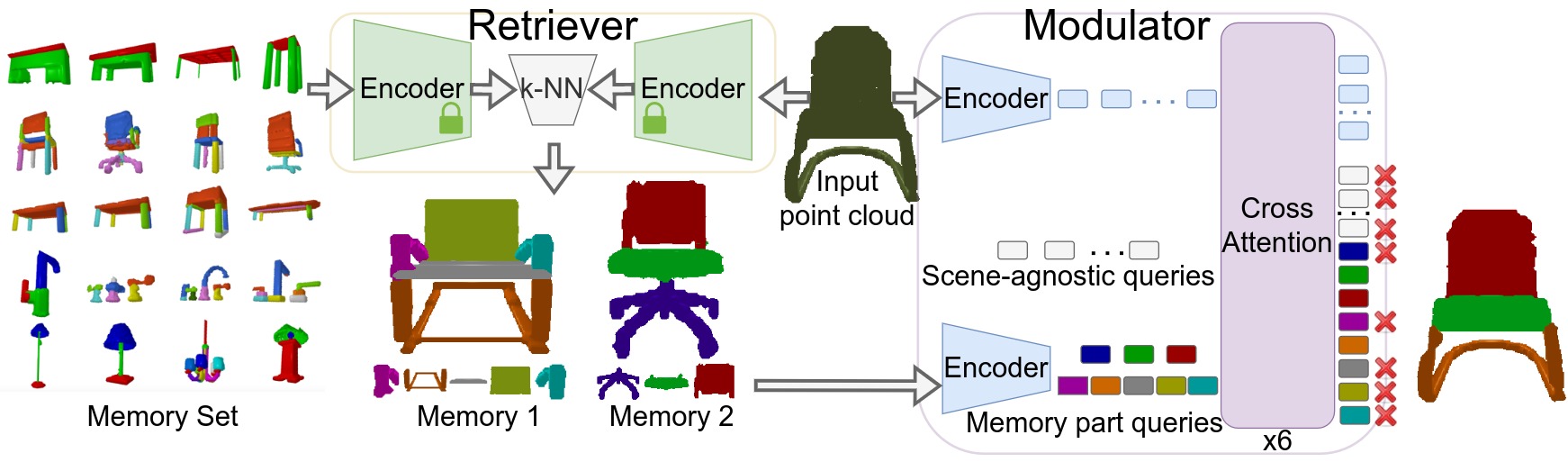}
  \caption{ 
  \textbf{Architecture for \model{}.} \model{} are comprised of   retriever and  modulator sub-networks.  
  In the retriever, labelled memories and the (unlabelled) input point cloud are  separately encoded into feature embeddings  and the top-$k$ most similar memories to the present input are retrieved. 
  In the  modulator,  each retrieved memory is encoded into a set of part feature embeddings and initializes a query that is akin to a slot to be ``filled" with the analogous part entity in the present scene. These queries are appended to a set of learnable parametric  scene-agnostic queries. The modulator contextualizes the queries with the input point cloud through iterative self and cross-attention operations that also update the point features of the input. 
  When a memory part query decodes a part in the input point cloud, we say the two parts are put into  correspondence by the model. We color them with the same color to visually indicate this correspondence.
  }
  \label{fig:architecture}
\end{figure}

The architecture of \model{} is illustrated in Figure \ref{fig:architecture}. 
\model{} are analogy-forming  transformer networks for part segmentation where queries that decode entities in the input scene are supplied by  retrieved part-encoded labelled scenes, as well as by the standard set   of scene-agnostic parametric queries of detection transformers \cite{Carion2020EndtoEndOD}. When a memory part query is used to decode a part segmentation in the input scene, we say the two parts, in the memory and the input, are put into correspondence. By using memory queries to decode parts in the input, our model forms analogies between detected part graphs in the input scene and part graphs in the modulating memories. Then, metadata, such as semantic labels, attached on memory queries automatically propagate to the detected parts.

\model{} are comprised of two main modules: (i) A retriever, that takes as input a 3D object point cloud and a memory repository of labelled 3D object point clouds,  and outputs  a set of relevant memories for the scene at hand, and  
(ii) a modulator, that jointly encodes the memories and the input  scene and predicts its 3D part segmentation.

\paragraph{Retriever}
The retriever has access to a  memory repository of labelled 3D object point clouds.  Each labelled training example is a memory in this  repository.  Examples labelled with different label granularities  constitute  separate memories. 
The retriever  encodes each memory example as well as the   input point cloud  into distinct normalized 1D feature encodings. 
The top-${k}$ memories are retrieved by computing an inner product between the input point cloud feature and the memory features. 

\paragraph{Modulator}  The modulator takes as input the retrieved memories and the unlabelled input point cloud and predicts part segments. 
The input scene is encoded into a set of 3D point features. Each memory scene is encoded into a set of 1D  part encodings, one for each annotated part in the memory, which we call memory part queries, in accordance to parametric queries used to decode objects in DETR \cite{Carion2020EndtoEndOD}. We additionally use scene-agnostic parametric queries, which are learnable slots that decode parts the memory cannot explain on its own. The input points,  memory part queries and scene-agnostic queries are contextualized via a set of cross and self-attention operations, that iteratively update all queries and point features. Each of the  contextualized queries predicts a segmentation mask proposal through inner-product with the contextualized point features, as well as an associated confidence score for the predicted part. 
These part segmentation proposals are matched to ground-truth part binary masks using  Hungarian matching  \cite{Carion2020EndtoEndOD}. 
For the mask proposals that are matched to a ground-truth mask, we compute the segmentation loss, which is a per-point binary cross-entropy loss 
\cite{Vu2022SoftGroupF3,Cheng2021PerPixelCI}. We also supervise the confidence score of all queries. 
  For implementation details, please see the Appendix, Section~\ref{psuedo_code}.

Our modulator network resembles detection and  segmentation transformers for 2D images \cite{Carion2020EndtoEndOD, DBLP:journals/corr/abs-2112-01527} where a set of learnable 1D vectors, termed parametric queries, iteratively cross-attend to input image features and self-attend among themselves to predict object segmentation masks in the input scene. The key difference between our modulator network and existing detection transformers are that \textbf{retrieved memory entity encodings are used as queries}, i.e., candidates for decoding parts in the input point cloud, alongside the standard set of scene-agnostic parametric queries, as shown in Figure \ref{fig:architecture}. Additional differences are that  we update the point features alongside the queries in the cross-attention layers, which we found helped performance. 

\subsection{Training}  
\model{} aim to learn to associate parts in the retrieved memory graphs with parts in the input point cloud. 
We train our model in two stages to facilitate this fine-grained association learning:

\begin{wrapfigure}{l}{0.25\textwidth}
     \begin{center}
 \includegraphics[width=1.0\linewidth]{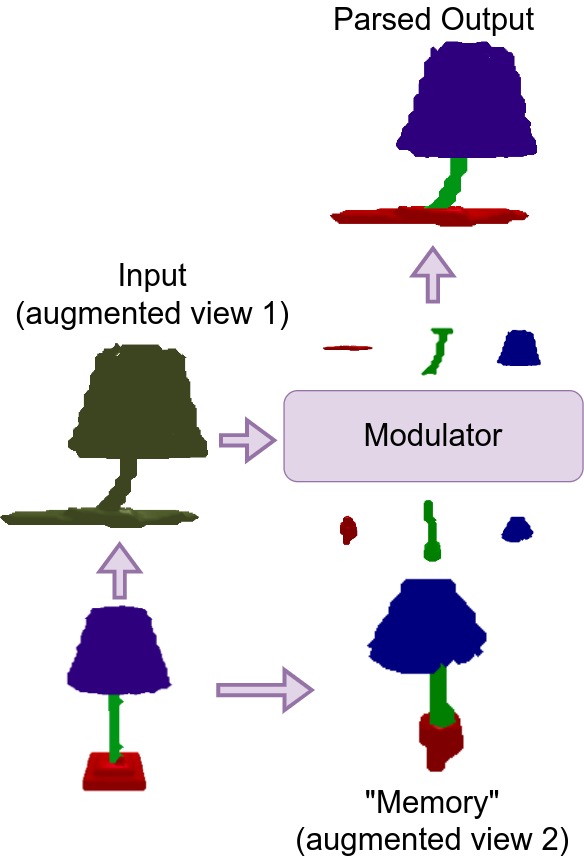}
     \end{center}
  \caption{\textbf{Within-scene pre-training. An augmented version of a scene is used as a memory for its parsing.} 
  }
  \label{fig:pretrain}
 \end{wrapfigure}

\textbf{1. Within-scene training:} 
We apply two distinct augmentations  (rotations and deformations \cite{kim2021pointwolf})  to each training scene, and use one as the input scene and the other as the modulating memory, as shown in Figure \ref{fig:pretrain}. In this  case, we \textbf{have access to ground-truth part associations between the parts of the memory  and the input scene}, which we use to supervise each memory part query to decode the corresponding  part in the input cloud, and we do not use Hungarian matching. We call this within-scene training since both the input and memory come from the same scene instance. 

\textbf{2. Cross-scene training:} 
During cross-scene training, the  modulating memories are  sampled from the top-$K$ retrieved memories per input scene. We show in our experimental section that \textbf{accurate memory query to input part associations emerge} during Hungarian matching in cross-scene training, 
thanks to  within-scene training.  This is important, as often we cannot have fine-grained annotations of structure correspondence across scene exemplars.

In \model{}, the retrieval process is not end-to-end differentiable with respect to the downstream scene parsing task. 
This is because i) we have no ground-truth annotations for retrieval, and ii) the retrieved  memories are contextualized with the input through  dense cross-attention operations 
which do not allow gradients to flow to the retriever's encoding parameters. 
We pre-train the encoder and  modulator parameters independently of the retriever using within-scene training. We then use the modulator's encoder weights as the retriever's frozen encoder (adding a parameter-free average pooling layer on top). In the Appendix, we include pseudo-code for within-scene training in Algorithm \ref{alg:within} and cross-scene training in Algorithm \ref{alg:cross}.

\newcommand{\cmark}{\ding{51}}%
\newcommand{\xmark}{\ding{55}}%

\section{Experiments}
We test \model{} on PartNet  \cite{Mo_2019_CVPR}, an established benchmark for   3D object segmentation. PartNet contains  3D object instances from multiple object categories, annotated with parts  in three different levels of granularity. 
We split PartNet object categories into base and novel categories. Our  base categories are Chair, Display, Storage Furniture, Bottle, Clock, Door, Earphone, Faucet, Knife, Lamp, Trash Can, Vase and our novel categories are Table, Bed,  Dishwasher and Refrigerator. 
We  consider two experimental paradigms: 
\textbf{1. Many-shot:}  For the exemplars of the base categories we consider the standard PartNet train/test splits. Our model and baselines are trained in the base category training sets, and tested  on segmenting instances of the base  categories in the test set. 
\textbf{2. Few-shot:} $K$ labelled examples from each novel category are given and the model is  tasked to segment new examples of these categories.  This tests few-shot adaptation. We consider $K=1$ and $K=5$. We aggregate results across multiple $K$-shot learning tasks (episodes).  
We test two versions of our model: i) \singleabl{}, which is \model{} with a single modulating memory, and ii)  \multiabl{}, with five modulating memories. Unless mentioned explicitly, \model{} will imply the single-memory model.
 
Our experiments  aim to answer the following questions:
\begin{enumerate}
\item How do \model{} compare against parametric-alone state-of-the-art models  in  many-shot and few-shot 3D object segmentation? 
\item How well do \model{} adapt few-shot via memory expansion, without any weight update?
\item  How  do \model{} compare against alternative memory-augmented networks  
where retrieved memory part encodings are attended to but not used as slots for  decoding  parts? 
\item  How well do \model{} learn part-based  associations across scenes without part association ground-truth or  semantic  label supervision? 
\end{enumerate}

\paragraph{Evaluation metrics}
We use the Adjusted Random Index (ARI) as our label-agnostic segmentation quality metric ~\citep{Rand1971ObjectiveCF}, which is a clustering score ranging from $-1$ (worst) to $1$ (best). ARI calculates the similarity between two point clusterings while being invariant to the order of the cluster centers. We compute $100 \times $ ARI using the publicly available implementation of ~\cite{multiobjectdatasets19}. 
We use mean Average Precision (mAP) per part \cite{Mo_2019_CVPR} for semantic part instance segmentation and mean intersection over union (mIoU) for 3D point semantic segmentation.

\paragraph{Baselines}
We compare our model to existing models in the literature. We further compare against strong parametric baseline models we develop. Our parametric baselines  already  outperform all  existing works  in the many-shot settings.  
We consider the following baseline models: 
\begin{itemize}[leftmargin=*]
\item \partnetbl{} of \cite{Mo_2019_CVPR} is a 3D segmentation network with the same backbone as our model. This model implements ``queries" as a fixed number of sets of MLPs that operate over the encoded input point cloud. Each set contains one MLP for per-pixel confidence and one for part confidence. The same losses used by our model are used to supervise this baseline.
\item \detrbl{}  is a 3D segmentation transformer network with the same backbone as our model and similar segmentation prediction heads and losses, but without any memory retrieval, akin to the 3D equivalent of a state-of-the-art 2D image segmentor  \cite{Carion2020EndtoEndOD, https://doi.org/10.48550/arxiv.2112.09131}. We update the point features  in the decoder layers, same as in our  model, which  we found to boost performance. Contrary to \detrblnospace, \model{} attend to external memories. 
 \item  \protobl{} is an adaptation of the episodic prototypical networks for image classification of \cite{NIPS2017_cb8da676} to the task of 3D object part segmentation. Specifically, given a set of N labelled point clouds, we form average feature vectors for each semantic labelled part and use them as queries to segment points into corresponding part masks through inner-product decoding. Contrary to \protobl{}, \model{} contextualize the memory queries and the input scene. \protobl{} require a retriever that knows the category label of the input scene, so it has access to privileged information. Without this assumption, we got very low performance from this model. 
\end{itemize}

\paragraph{Ablations}
We  compare our model to the following variants and ablative versions:
\begin{itemize}[leftmargin=*]
 \item \memtransfabl{} (Retrieval-DETR3D) is a variant of \model{} that attends to 
retrieved part memory encodings but does not use them to decode parts. Instead, all parts are decoded from the scene-agnostic queries. 
Different from \model{}, analogies cannot emerge between a memory and the input scene since this model does not represent such correspondence explicitly, but only implicitly, in the attention  operations. 
\item  \nopretrain{}  is our  \singleabl{} model without any within-scene (pre)training. 

\item \modelgtret{} is our  \singleabl{} model with a privileged retriever that has  access to the ground-truth category label of the input  and only retrieves memories of the same category label. 

\end{itemize}

\subsection{Many and few-shot 3D object  segmentation}\label{sec:4.1}
The PartNet benchmark provides three levels of segmentation annotations per object instance where level 3 is the most fine-grained. 
We train and test our model and baselines on all three  levels.  
We use a learnable level embedding as additional input for our baselines \partnetbl{} and \detrbl{},  as is usually the case in multi-task models \cite{DBLP:journals/corr/abs-2202-02005}. 
In the many-shot setting, we train our model and baselines jointly across all base categories  and test them across all of them as well, using the standard PartNet train/test splits. 
For \model{} and \memtransfabl{},  all examples in the training set become part of their memory repository. 
In the few-shot setting, \partnetbl{} and \detrbl{} adapt by weight finetuning on the $K$-shot task. 
\modelnospace{} and \memtransfabl{} adapt in two ways: i) by expanding the memory of the model with the novel $K$-shot support examples, and ii) by further adapting the weights via fine-tuning to segment the $K$ examples. 
In this case,  the memory set is  only the novel labelled support set instances.  \textbf{The retriever does not have access to the object category information in any of the many-shot or few-shot settings} unless explicitly stated so.

\begin{table}[]
\resizebox{\columnwidth}{!}{%
\begin{tabular}{|l|c|cc|c|}
\hline
\multirow{2}{*}{Method} & \multirow{2}{*}{Fine-tuned?} & \multicolumn{2}{c|}{Novel Categories} & Base Categories \\ 
 &  & 1-shot ARI ($\uparrow$) & 5-shot ARI ($\uparrow$) & Many shot ARI ($\uparrow$)\\ \hline
 
\multirow{2}{*}{\partnetbl{} } & \xmark & 26.1 & 26.1  & 54.3  \\
 & \cmark & $22.0 \pm 0.90$ & $25.9 \pm 0.87$ &  - \\ \hline
 
\multirow{2}{*}{\detrbl{}} & \xmark & 30.4 & 30.4 & \textbf{74.3}  \\
 & \cmark & $39.4 \pm 1.44$ & $52.7 \pm 1.44$ &  - \\ \hline
 
\multirow{2}{*}{\begin{tabular}[c]{@{}c@{}}\memtransfabl{}\end{tabular}} & \xmark & $38.2 \pm 1.79$ & $46.8 \pm 0.66$ & \textbf{74.3} \\ 
 & \cmark & $46.5 \pm 2.61$ & $55.6 \pm 1.82$ & - \\ \hline

\multirow{2}{*}{\begin{tabular}[c]{@{}c@{}}\singleabl{} \end{tabular}} & \xmark & $\textbf{49.0} \pm \textbf{0.80}$ & $52.0 \pm 1.11$ & $72.5$  \\ 
 & \cmark & $\textbf{50.6} \pm \textbf{2.72}$ & $\textbf{57.0} \pm \textbf{1.33}$  & - \\ \hline
 
\multirow{2}{*}{\begin{tabular}[c]{@{}c@{}}\multiabl{} \end{tabular}} & \xmark & - & $\textbf{52.1} \pm \textbf{0.75}$ & 74.2 \\
 & \cmark & - & $56.4 \pm 1.81$ & -  \\ \hline \hline
 
 \multirow{2}{*}{\begin{tabular}[c]{@{}c@{}}\modelgtret{}{} \end{tabular}} & \xmark & $51.2 \pm 0.96$ & $53.8 \pm 1.03$ & {75.6} \\
 & \cmark & ${52.6} \pm {2.96}$ & $58.3 \pm 1.36$ & -  \\ \hline
 
\end{tabular}
}
\caption{
\textbf{Semantics-free 3D part segmentation performance} on the PartNet benchmark. 
Without any fine-tuning, \model{} outperform \detrbl{} by more than 20\% in the few-shot setup.  Even upon fine-tuning, \model{} outperform \detrbl{} by 4.3\% ARI. 
} 
\label{tab:results}
\end{table}

\subsubsection{Semantics-free part instance segmentation}
We train all models and baselines for object part segmentation objectives, without any semantic label information, as described in Section \ref{sec:prelim}.    
We show quantitative  many-shot and few-shot part segmentation results in Table \ref{tab:results}. For the few-shot setting, we show both  fine-tuned and non-fine-tuned models. For the few-shot performance, we report mean and standard deviation over 10 tasks where we vary the $K$-shot support set. 
Our conclusions are as follows:  

 \textbf{(i)} \model{} dramatically outperform  \detrbl{} in few-shot part segmentation. While in the many-shot setting the two models have similar performance, when adapting few-shot to  novel categories, \model{} and all their variants dramatically outperform parametric alone \detrbl{}, both before and after fine-tuning.  
 
\textbf{(ii)} Analogical networks can adapt few-shot  simply by memory expansion,  without weight updates.  Indeed, the 5-shot performance of our model is close before and after fine-tuning in the novel categories (52.0\% versus 57.0\% ARI). 

\textbf{(iii)} \multiabl{} outperform the single-memory version in many-shot learning with on par few-shot performance. 

\textbf{(iv)} \memtransfabl{} adapt few-shot  better than  \detrbl{}. 
Still, before weight fine-tuning in the few-shot test set, this memory-augmented variant exhibits   worse performance than our single-memory model  (46.8\% versus 52.0\% ARI). 

\textbf{(v)} A  retriever that better recognizes object categories would provide a performance boost, especially in the many-shot setting. 

\subsubsection{Emergent cross-scene correspondence}
\model{} can label the parts they segment with semantic categories by propagating semantic labels of the memory parts that were used to decode the input parts. We measure  semantic instance segmentation performance for our model and baselines in Table \ref{tab:result_segmentation}. 
All variants of \model{} are trained \textit{without any semantic labelling objectives. }
On the contrary, we  train the baselines \detrbl{} and \protobl{} for \textit{both  object part segmentation and part labelling.} \model{} predict semantic part labels only via propagation from memory part queries that decode parts in the scene, and does not produce any semantic labels for parts decoded by scene-agnostic queries, so by default it will make a mistake each time a parametric query is used.  We found more than 80\% of parts are decoded by memory part queries on average, while for \multiabl{} this ratio is 98\%.  
For the few-shot settings in Table \ref{tab:result_segmentation}, all models are fine-tuned on the few given examples.   
Similar to our model, \protobl{} propagate semantic labels of the prototypical part features. We evaluate \protobl{} only for semantic segmentation since it cannot easily produce instance segments:  if multiple part instances share the same semantic label, this model assumes they belong  to the same semantic prototype. 
Our conclusions are as follows:

\textbf{(i)} \model{}  show very competitive semantic and instance segmentation accuracy 
via label propagation through memory part queries, despite having never seen semantic labels at training time. This shows \textbf{our model learns cross-scene part associations without any semantic information.}

\textbf{(ii)} \singleabl{} \texttt{w/o within-scene} has much worse semantic segmentation performance which suggests  \textbf{within-scene training much helps cross-scene associations to emerge without semantic information.}

\textbf{(iii)} \protobl{} achieves high 1-shot performance but does not scale with more data and is unable to handle both few-shot and many-shot settings efficiently.

\textbf{(iv)} A retriever that better recognizes object categories  boosts performance of \model{} over \detrbl{} in the few-shot settings.

\begin{table}[t!]
\resizebox{\columnwidth}{!}{%
\begin{tabular}{|l|cc|cc|cc|}
\hline
\multicolumn{1}{|c|}{\multirow{2}{*}{Method}}                                                                   & \multicolumn{2}{c}{\begin{tabular}[c]{@{}c@{}}Novel Categories\\ 1-shot\end{tabular}} & \multicolumn{2}{c}{\begin{tabular}[c]{@{}c@{}}Novel Categories\\ 5-shot\end{tabular}} & \multicolumn{2}{c|}{\begin{tabular}[c]{@{}c@{}}Base Categories\\ Many-shot\end{tabular}} \\ \cline{2-7} 

\multicolumn{1}{|c|}{} & mIoU & mAP & mIoU & mAP & mIoU & mAP \\ \hline
\begin{tabular}[c]{@{}l@{}}\detrbl{}$^*$ \end{tabular}  & $21.5$ & $18.3$ & $30.6$ & $27.5$ & $55.9$ & $53.6$ \\ \hline

\begin{tabular}[c]{@{}l@{}} \singleabl{} \texttt{w/o within-scene}\end{tabular} & $5.0$ & $3.3$ & $4.7$ & $3.9$ & $7.8$ & $6.2$ \\ \hline

\begin{tabular}[c]{@{}l@{}} \singleabl{}{}\end{tabular} & $20.4$ & $18.2$ & $26.0$ & $25.0$ & $44.3$ & $42.0$ \\ \hline

\begin{tabular}[c]{@{}l@{}} \multiabl{}{}\end{tabular} & - & - & $27.8$ & $25.8$ & $49.2$ & $47.9$ \\ \hline \hline

\begin{tabular}[c]{@{}l@{}}\protobl{}$^*$ \end{tabular} & $27.5$ & - & $29.0$ & - & $30.0$ & - \\ \hline

\begin{tabular}[c]{@{}l@{}} \modelgtret{} \end{tabular} & $26.2$ & $25.2$ & $30.2$ & $30.2$ & $50.6$ & $48.7$ \\ \hline
\end{tabular}}
\caption{
\textbf{Part Semantic and Part Instance Segmentation performance}  on the PartNet benchmark. $^*$ indicates training with semantic labels.
}
\label{tab:result_segmentation}
\end{table}

\begin{table}[t!]

\begin{tabular}{|l|c|c|c|}
\hline
\multirow{2}{*}{Method} & \multirow{2}{*}{Fine-tuned?} & Novel Categories & Chair \\ \cline{3-4} 
 &  & 5-shot ARI ($\uparrow$) & ARI ($\uparrow$) \\ \hline
 
 \multirow{2}{*}{\detrbl{} trained on ``Chair''} & \xmark &  \multicolumn{1}{c|}{$28.5$} & $76.0$ \\
 & \cmark & \multicolumn{1}{c|}{$47.7 \pm 2.31$} & - \\ \hline

\multirow{2}{*}{\detrbl{}} & \xmark &  \multicolumn{1}{c|}{$30.4$} & $75.7$ \\
 & \cmark & \multicolumn{1}{c|}{$52.7 \pm 1.44$} & - \\ \hline
 
 \multirow{2}{*}{\begin{tabular}[c]{@{}c@{}}\model{}   trained on ``Chair'' \end{tabular}} & \xmark & \multicolumn{1}{c|}{$33.8 \pm 0.89$} & $\textbf{76.5}$ \\
 & \cmark & \multicolumn{1}{c|}{$50.4 \pm 1.60$} & - \\ \hline
 
 \multirow{2}{*}{\begin{tabular}[c]{@{}c@{}}\model{} \end{tabular}} & \xmark & $\textbf{52.0} \pm \textbf{1.11}$ & $76.3$ \\
 & \cmark & $\textbf{57.0} \pm \textbf{1.33}$ & - \\ \hline

\end{tabular}
\caption{ \textbf{Few-shot learning for single-category  and multi-category trained models.} \model{} learn better few-shot  when trained across all categories, while  \detrbl{}   does not improve its few shot learning performance when trained across more categories.}
\label{tab:results_single_cat}
\end{table}

\subsubsection{Multi-category training helps few-shot adaptation in \model{}} \label{multic} 
We compare our model and baselines on their ability to improve  few-shot learning performance with more diverse training data in Table \ref{tab:results_single_cat}. We train each model under two setups: i) training only on instances of ``Chair'', which is the most common category---approximately 40\% of our training examples fall into this category---(refer to Table \ref{tab:freq} for statistics of the training data distribution) and, ii) 
training on ``all" categories. We test the performance of each model on 5-shot learning. 
Our conclusions are as follows: \textbf{(i) Models trained on a single category usually fail to few-shot generalize to other classes} with or w/o fine-tuning, despite their strong performance on the training class. 
\textbf{(ii)} \detrbl{} does not improve in its ability to adapt few-shot with more diverse training data, in contrast to \model{}.

In the Appendix, we show the effect of different retrieval mechanisms (\ref{supp_rtvr}, \ref{rtrvr2}) 
and qualitative results on more benchmarks (\ref{scannet}).

\subsection{Limitations - Discussion}
A set of future directions that are necessary for \model{} to scale beyond segmentation of single-object scenes are the following: 
\textbf{(i)} The retriever in \model{} operates currently over whole object memories and is not end-to-end differentiable with respect to the downstream task. Sub-object part-centric memory representations would permit fine-grained retrieval of visual memory scenes. We further plan to explore alternative supervision for the retriever module inspired by works in the language domain \cite{https://doi.org/10.48550/arxiv.2208.03299,izacard:hal-03463398}. 
\textbf{(ii)}  Scaling \model{} to segmentation of complete,  multi-object 3D  scenes in realistic home environments  requires scaling up the size of memory collection.  It would further necessitate bootstrapping fine-grained object part annotations,  missing from 3D scene datasets \cite{dai2017scannet}, by transferring knowledge of object part compositions from PartNet. Such semi-supervised fine-grained scene parsing is an exciting avenue of future work.
\textbf{(iii)} So far we have assumed the input to \model{} to be a complete 3D point cloud. However, this is hardly ever the case in reality. Humans and machines  need to make sense of single-view, incomplete and noisy observations. Extending \model{} with generative heads that not only detect analogous parts in the input, but also inpaint or complete missing parts, is a direct avenue for future work.

\section{Conclusion}

We  presented \modelnospace, a  semi-parametric model for associative 3D visual parsing that  
puts forward an analogical paradigm of corresponding  input scenes to compositions and modifications of  memory scenes and their  labelled parts, instead of   mapping  scenes to segments directly. 
One-shot, few-shot or many-shot learning are treated uniformly in \model{}, by conditioning to the appropriate set of memories, whether taken from a single, few or many memory exemplars, and inferring analogous parses. We showed \model{} outperform SOTA parametric and meta-learning baselines in  few-shot 3D object parsing. We further showed  correspondences emerge across scenes without semantic supervision, as a by-product of the analogical inductive bias and our within-scene  augmentation training. 
In his seminal work ``the proactive brain" ~\citep{Bar2007-BARTPB-2}, Moshe Bar argues for the importance of analogies and associations in human reasoning---highlighting how associations of novel inputs to analogous representations in memory can drive perceptual inference. \model{} operationalize these insights in a retrieval-augmented  3D parsing framework, with analogy formation between retrieved memory graphs and  input scene segmentations.
\section*{Reproducibility Statement}

To ensure the reproducibility of the empirical results, we include a pseudo-code of the main model components and training pipeline in the Appendix. We have made our code publicly available on GitHub.

\subsubsection*{Acknowledgments} 
 This work is supported by Sony AI, a DARPA Young Investigator Award, an NSF CAREER award, an AFOSR Young Investigator Award, a Verisk AI Faculty Research Award, and DARPA Machine Common Sense.

{
\bibliographystyle{iclr2023_conference}
\bibliography{darpa,iclr2022,refs}
}
\newpage
\section{Appendix}
Our Appendix is organized as follows: In Section \ref{psuedo_code}, we provide implementation details and pseudo-code for training \model{}. In Section \ref{supp_rtvr} we ablate \model{}' performance under varying memory retrieval schemes in $5$-shot setting and show qualitative results for the retriever in Section \ref{rtrvr2}. We show more results on noisy point clouds (ScanObjectNN \cite{Uy2019RevisitingPC} dataset) in \ref{scannet}. We provide extensive qualitative visual object parsing results for single and multi-memory variants of \model{} in section \ref{parse_fig}. Lastly, we discuss additional related literature in Section \ref{sec:additional_rel_work}.

\subsection{Implementation Details and Training Pseudo Code}
\label{psuedo_code}

The modulator  encodes the input scene $S$ 
and each retrieved memory scene $M$ into a set of  3D point features using PointNet++ backbone.  
We encode positional information using rotary 3D positional encodings \cite{Su2021RoFormerET,Li2021LepardLP}, which have the property that $P(x)^TP(y) = P(y-x)$, where $P$ the positional encoding function and $x, y$ two 3D points. These embeddings are multiplied with queries and keys in the attention operations, thus making attention translation-invariant. For memory queries, we use the centroid of the corresponding memory part to compute positional encodings. For scene-agnostic queries, we use the center of the input object.

Each labelled part $p$ in each memory $M$ is encoded into an 1D feature vector $f_p^M$ by average pooling  its point features. 
Next, the queries (memory and scene-agnostic combined) self-attend and cross-attend to the input point features and update themselves; point features self-attend and cross-attend to queries to also update themselves. This input feature update is a difference from existing detection transformers, where only the queries are updated. We consider $6$ layers of self and  cross-attention.

Lastly, we upsample the point features to the original resolution using convolutional layers  \cite{Qi2017PointNetDH} and compute an inner product between each query and point feature to compute the segmentation mask for each query.  
The output of the modulator is a set of $N_q$ segmentation mask proposals and corresponding confidence scores, where $N_q$ is the total number of queries. At training time, these proposals are matched to ground-truth instance binary masks using the Hungarian matching algorithm \cite{Carion2020EndtoEndOD}. For the proposals that are matched to a ground-truth instance, we compute the segmentation loss, which is a per-point binary cross-entropy loss 
\cite{Vu2022SoftGroupF3,Cheng2021PerPixelCI}. We also supervise the confidence score of each query, similar to \cite{Carion2020EndtoEndOD}. The target labels are $1$ for the proposals matched with a ground-truth part and $0$ for non-matched. We found it beneficial to apply these losses after every cross-attention layer in the modulator. At test time, we multiply the per point mask occupancy probability with tiled confidence scores to get a $N_p \times N_q$ tensor ($N_p$ is the number of points); then each point is assigned to the highest scoring query by computing inner product and taking per-point argmax over the queries. 
The modulator's weights are  trained with within-scene and cross-scene training where the modulating memories are  sampled from the top-k retrieved memories. During cross-scene training, we further co-train with within-scene training data.

For both stages of training (i.e. within-scene correspondence pre-training and cross-scene training), we use AdamW optimizer \citep{loshchilov2017adamw} with an initial learning rate of $2\mathrm{e}{-4}$ and batch size of $16$. We train the model for $100$ epochs within-scene and $60$ cross-scene. For few-shot fine-tuning/evaluation, we use AdamW optimizer with an initial learning rate of $3\mathrm{e}{-5}$ and batch size of $8$. We fine-tune for $90$ epochs and we report the performance across $10$ different episodes, where each episode has a different set of K support samples. We describe \model{}' training details in pseudo-code for within (Algorithm~\ref{alg:within}) and cross-scene training (Algorithm~\ref{alg:cross}) respectively.

For our \detrbl{} baseline we use same hyperparameters and train for $250$ epochs. Training takes approximately 15 and 20 minutes per epoch on a single NVIDIA A100 gpu for \detrbl{} and \model{} respectively. For the multi-memory model we reduce the batch size to 8 and the learning rate to $1\mathrm{e}{-4}$. Each epoch takes around 50 minutes.

\begin{algorithm}[t]
  \caption{Pseudo code for within-scene correspondence pre-training of \model{} }
  \label{alg:within}
    \definecolor{codeblue}{rgb}{0.25,0.5,0.5}
    \definecolor{codekw}{rgb}{0.85, 0.18, 0.50}
    \newcommand{\algofontsize}{8.0pt}
    \lstset{
      backgroundcolor=\color{white},
      basicstyle=\fontsize{\algofontsize}{\algofontsize}\ttfamily\selectfont,
      columns=fullflexible,
      breaklines=true,
      captionpos=b,
      commentstyle=\fontsize{\algofontsize}{\algofontsize}\color{codeblue},
      keywordstyle=\fontsize{\algofontsize}{\algofontsize}\color{black},
    }
\begin{lstlisting}[language=python]
# S: input point cloud, M: memory point cloud, Np: numbers of points in S or M, N: sub-sampled points, C: number of feature channels, P: number of parts in M
# augment: a sequence of standard 3D point cloud augmentations
# pc_encoder: point cloud encoder
# Xp(M): ground-truth label assignment of points in parts, copied directly from the memory
# part_encoder: Computes the part features using mean pooling
# pos_encode: Adds positional encoding
# upsampler: Upsamples point cloud
# Segmentation_Loss: Cross entropy loss to assign each point to the Hungarian matched query.

for S in dataloader: # load a batch with B samples
    M = S  # the memory is the un-augmented version
    S = augment(S)  # the input is augmented
    # S : B x Np x 3 and  M:  B x Np x 3
    # Compute point features
    F^S = pc_encoder(S) # B x N x C
    F^M = pc_encoder(M) # B x N x C
    
    # Initialize memory part queries
    f^M = part_encoder(F^M) # B x P x C
    
    # Compute positional embedding
    x_pos = pos_encode(F^S)
    y_pos = pos_encode(f^M)  # B x P x C
    
    Loss = 0    
    # Do multiple layers of modulation using Self-Attn and Cross-Attn
    for layer in num_layers:
        x = Cross-Attn(x, y, x_pos, y_pos) # B x N x C
        y = Cross-Attn(y, x, y_pos, x_pos) # B x P x C
        x = Self-Attn(x, x_pos) # B x N x C
        y = Self-Attn(y, y_pos) # B x P x C
        
        X = upsampler(x) # B x Np x C
        point_query_similarity = matmul(normalize(X), normalize(y.T)) # B x Np x P
        
        Loss += Segmentation_Loss(argmax(point_query_similarity, -1), Xp(M))
        
    
    # optimizer step
    loss.backward()
    optimizer.step()
\end{lstlisting}
\end{algorithm}

\begin{algorithm}[t]
  \caption{Pseudo code for cross-scene training of \model{} }
  \label{alg:cross}
    \definecolor{codeblue}{rgb}{0.25,0.5,0.5}
    \definecolor{codekw}{rgb}{0.85, 0.18, 0.50}
    \newcommand{\algofontsize}{8.0pt}
    \lstset{
      backgroundcolor=\color{white},
      basicstyle=\fontsize{\algofontsize}{\algofontsize}\ttfamily\selectfont,
      columns=fullflexible,
      breaklines=true,
      captionpos=b,
      commentstyle=\fontsize{\algofontsize}{\algofontsize}\color{codeblue},
      keywordstyle=\fontsize{\algofontsize}{\algofontsize}\color{black},
    }
\begin{lstlisting}[language=python]
# S: input point cloud, M: retrieved memory point cloud, Np: numbers of points in S or M, N: sub-sampled points, C: number of feature channels, P: number of parts in M, target_classes: semantic classes of ground-truth parts of S
# Q: number of learnable scene-agnostic queries
# pc_encoder: point cloud encoder
# Xp_Hungarian: Hungarian matched label assignment of points in S to queries
# yp_Hungarian: Hungarian matched label assignment of queries to GT parts in S
# matched: indices of queries that have been matched to a ground-truth part
# part_encoder: Computes the part features using mean pooling
# pos_encode: Adds positional encoding
# upsampler: Upsamples point cloud
# Objectness_Loss: Binary cross entropy loss to decide which queries (scene-agnostic+learnable) would be responsible for decoding parts
# Segmentation_Loss: Cross entropy loss to assign each point to the hungarian matched query.
# Semantic_Loss: Cross entropy loss to map hungarian matched queries to semantic classes

for S,M in dataloader: # load a batch with B samples
    # S : B x Np x 3 and  M:  B x Np x 3
    # Compute point features
    F^S = pc_encoder(S) # B x N x C
    F^M = pc_encoder(M) # B x N x C
    
    # Initialize memory part queries
    f^M = part_encoder(F^M) # B x P x C
    
    # Compute positional embedding
    x_pos = pos_encode(F^S)
    y = pos_encode(Concatenate(f^M, scene_agnostic_queries))  # B x (P + Q) x C
    
    Loss = 0    
    # Do multiple layers of modulation using Self-Attn and Cross-Attn
    for layer in num_layers:
        x = Cross-Attn(x, y, x_pos, y_pos) # B x N x C
        y = Cross-Attn(y, x, y_pos, x_pos) # B x P x C
        x = Self-Attn(x) # B x N x C
        y = Self-Attn(y) # B x P x C
        
        X = upsampler(x) # B x Np x C
        point_query_similarity = matmul(normalize(X), normalize(y.T)) # B x Np x (P + Q)
        
        Loss += Segmentation_Loss(armax(point_query_similarity, -1), Xp_Hungarian) + Objectness_Loss(y, yp_Hungarian) + Semantic_Loss(y[matched], target_classes)
        
    
    # optimizer step
    loss.backward()
    optimizer.step()
\end{lstlisting}
\end{algorithm}

\begin{table}[t!]
\resizebox{\columnwidth}{!}{%
\begin{tabular}{|c|c|l|c|}
\hline
Method & Fine-tuned? & Modulating Memory & \begin{tabular}[c]{@{}c@{}}Novel Category:\\ 5-shot ARI ($\uparrow$) \end{tabular} \\ \hline
\multirow{5}{*}{\singleabl{} } & \multirow{5}{*}{\xmark} & Random category-constr. & $48.5 \pm 0.76$ \\ 
 &  & Retriever category-constr. & $53.8 \pm 1.03$ \\ 
 &  & Oracle &  $61.9 \pm 1.22$\\ \hline
\multirow{5}{*}{\singleabl{}} & \multirow{5}{*}{\cmark} & Random category-constr. & $53.5 \pm 1.28$ \\ 
 &  & Retriever category-constr. &  $58.3 \pm 1.36$ \\ 
 &  & Oracle &  $62.3 \pm 0.66$\\ \hline
\end{tabular}}
\caption{\textbf{Ablations on ARI segmentation performance under varying retrieval schemes for 5-shot on 4 novel categories}.}
\label{tab:ablations}
\end{table}

\subsection{Performance under Varying Retrieval Schemes}
\label{supp_rtvr}
In the few-shot setting, we evaluate the performance of \model{} under varying memory retrieval schemes in Table \ref{tab:ablations}. We compare against a hypothetical \textit{oracle} retriever that can fetch the most helpful (in terms of resulting ARI) memory for each input. All examined retrievers are category-constrained, i.e., they have access to the object category of the input and retrieve an object of the same category. Our conclusions are as follows: \textbf{(i) \model{} with an oracle memory retriever perform better than \model{} using memories retrieved by  the retriever}. This suggests that better training of our retriever or exploring its co-training with the rest of our model could have significant impact in improving its performance. \textbf{(ii) Considering any of the 5-shot exemplars randomly does worse than using memories retrieved by our retriever.}

\begin{table}[t!]
\centering
\resizebox{\columnwidth}{!}{%
\begin{tabular}{|c|c|c|c|c|c|c|c|c|c|c|c|c|c|c|c|}
\hline
\multicolumn{12}{|c|}{Base Categories} & \multicolumn{4}{c|}{Novel Categories} \\ \hline
 Chair & Display & \begin{tabular}[c]{@{}c@{}}Storage\\ Furniture\end{tabular} & Bottle & Clock & Door & \begin{tabular}[c]{@{}c@{}}Ear\\ phone\end{tabular} & Faucet & Knife & Lamp & \begin{tabular}[c]{@{}c@{}}Trash\\ Can\end{tabular} & Vase & Table & Bed & Dishwasher & Refrigerator\\ \hline

6323 & 928 & 2269 & 436 & 554 & 225 & 228 & 648 & 327 & 2207 & 321 & 1076 & 8218 & 194 & 181 & 187 \\\hline 
\end{tabular}}
\caption{
{\textbf{Number of samples per category in the PartNet dataset \cite{Mo_2019_CVPR}.} Note that each sample has annotations for three levels of segmentation granularity.}
}
\label{tab:freq}
\end{table}

\subsection{Qualitative Performance of the Retriever} \label{rtrvr2}
\begin{figure}[h!]
  \centering
  \includegraphics[width=0.9\linewidth]{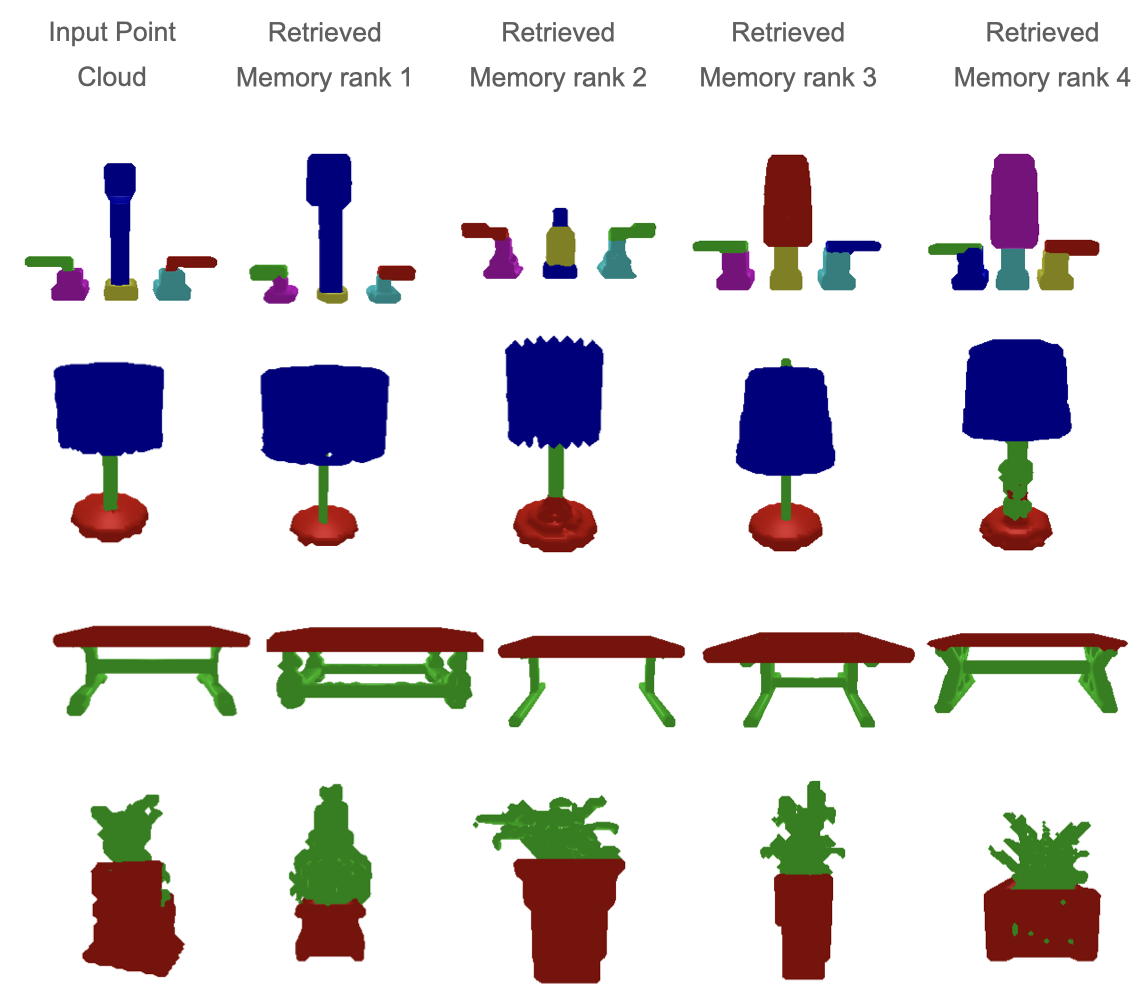}
    \includegraphics[width=0.9\linewidth]{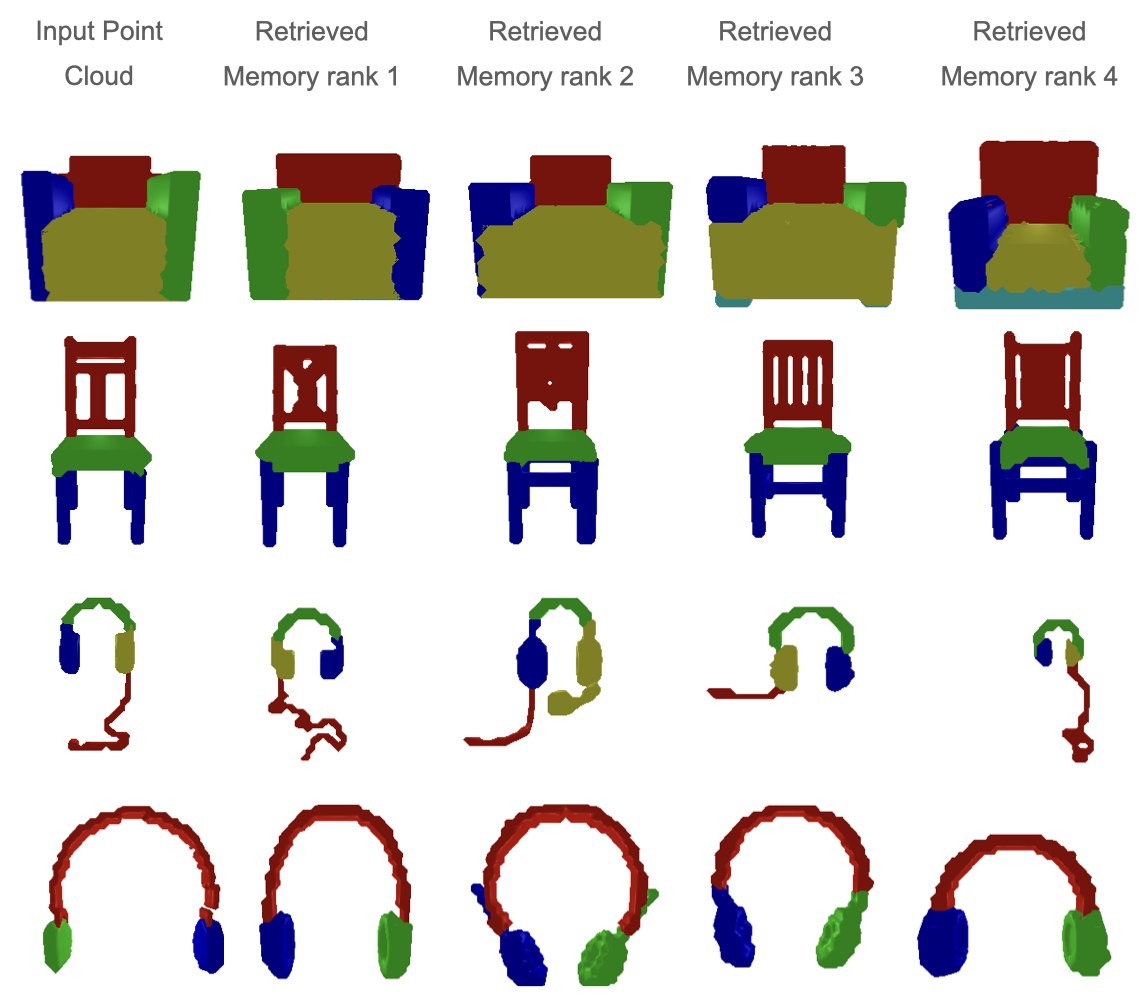}
  \caption{Top-$4$ retrieved results for each input point cloud. Examples from base classes of PartNet \cite{Mo_2019_CVPR} dataset. Note that instances of the same category can retrieve different memories, focusing on structural similarity and not only semantic.} 
  \label{fig:viz_retriever_1}
\end{figure}

\begin{figure}[h!]
  \centering
    \includegraphics[width=1.0\linewidth]{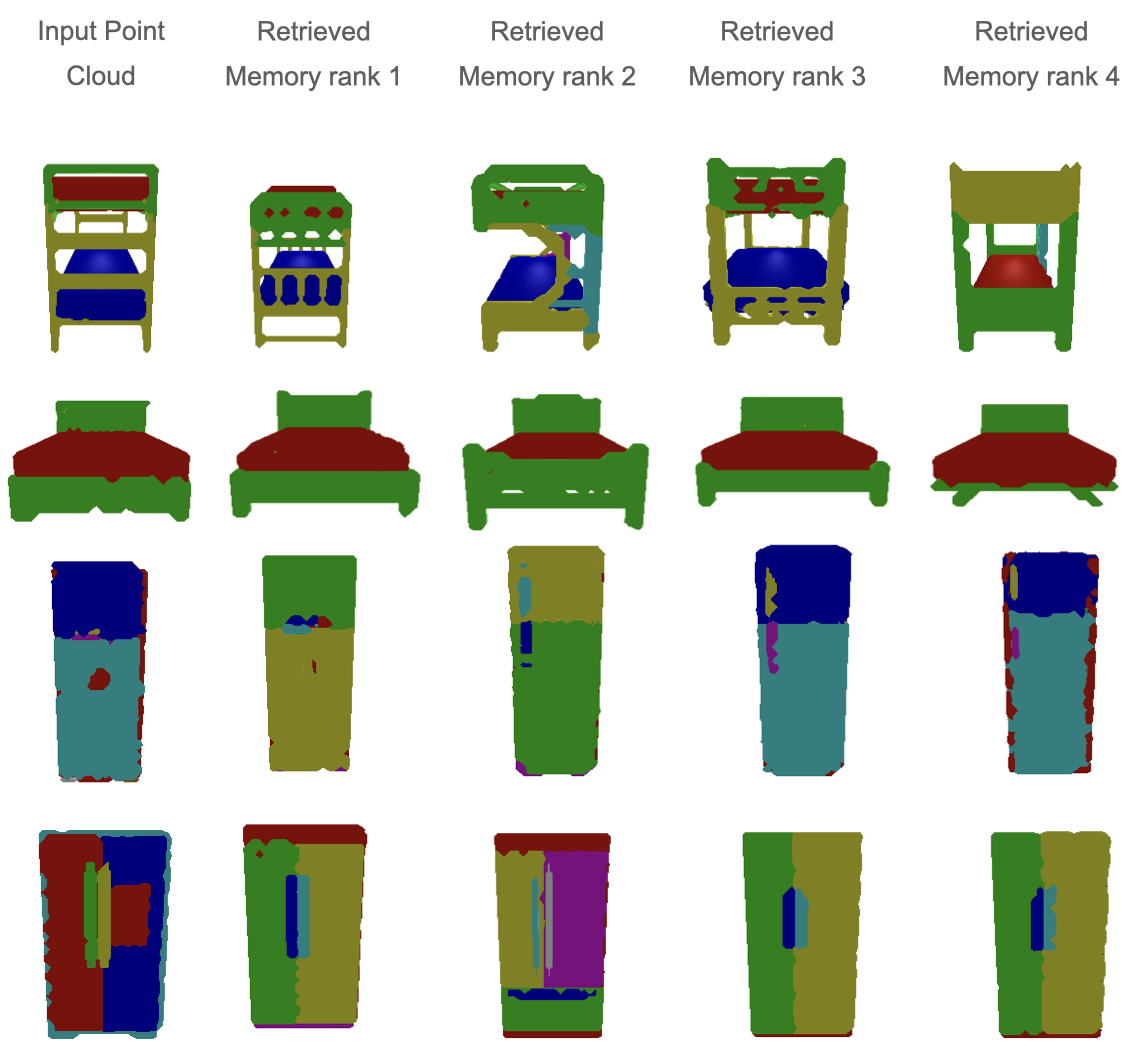}
  \caption{Top-$4$ retrieved results for each input point cloud. Examples from novel classes of PartNet \cite{Mo_2019_CVPR} dataset. Note that instances of the same category can retrieve different memories, focusing on structural similarity and not only semantic. This behavior generalizes to novel classes as well, even if the model has never seen such geometries before.} 
  \label{fig:viz_retriever_2}
\end{figure}

We show qualitative results of the retriever on multiple classes, both seen (Figure \ref{fig:viz_retriever_1}) during training and unseen (Figure \ref{fig:viz_retriever_2}). We observe the following: \textbf{(i) The retriever considers fine-grained object similarities and not only class information.} To illustrate this, we include two examples for the ``Chair" and ``Earphone" classes in Figure \ref{fig:viz_retriever_1}, as well as the ``Bed" and ``Refrigerator" classes in Figure \ref{fig:viz_retriever_2}. Different instances of the same category retrieve very different memories, that share both structural and semantic similarities with the respective input point cloud. \textbf{(ii) The retriever generalizes to novel classes, not seen during training}, as shown in Figure \ref{fig:viz_retriever_2}.

\subsection{Evaluation on ScanObjectNN Dataset \cite{Uy2019RevisitingPC}} \label{scannet}
\begin{figure}[h!]
  \centering
  \includegraphics[width=0.9\linewidth]{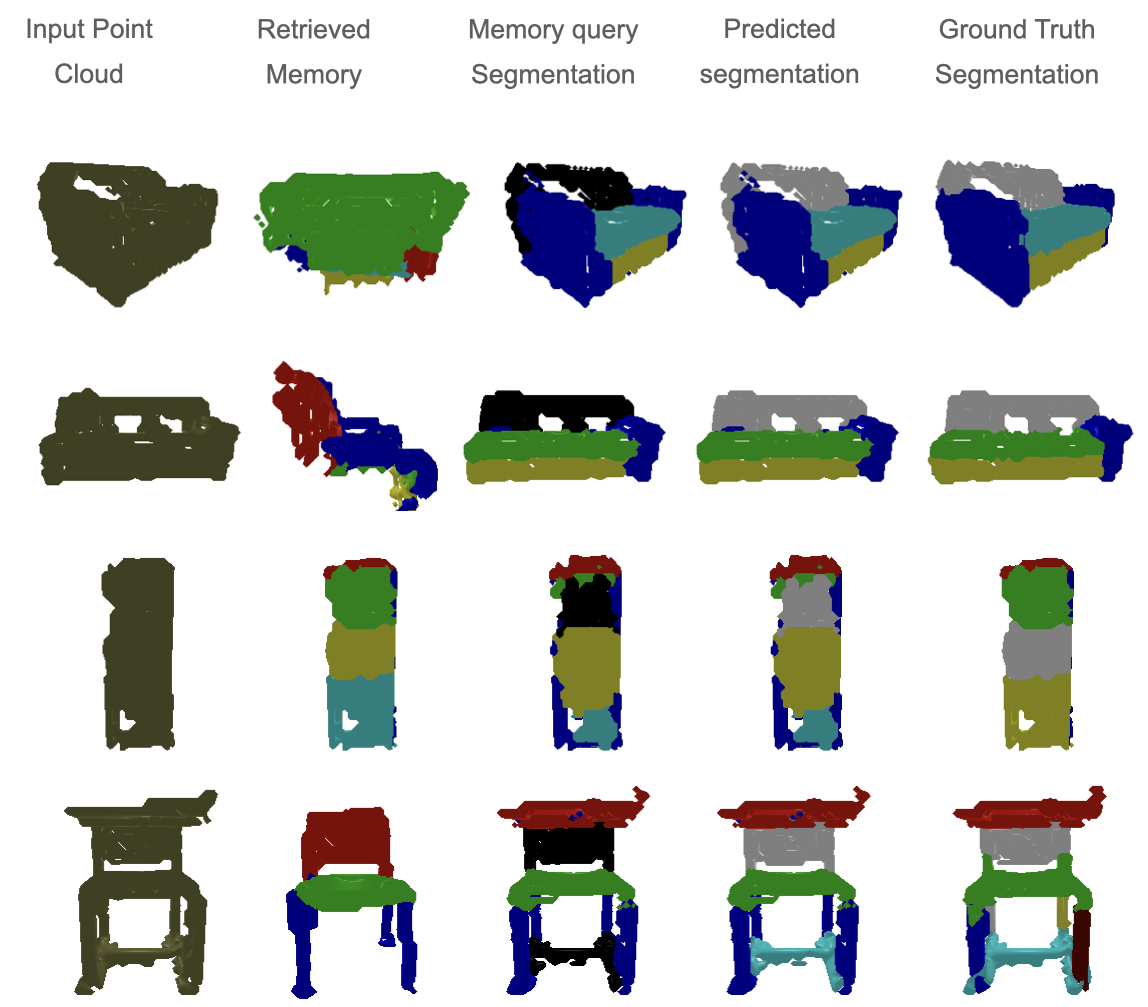}
    \includegraphics[width=0.9\linewidth]{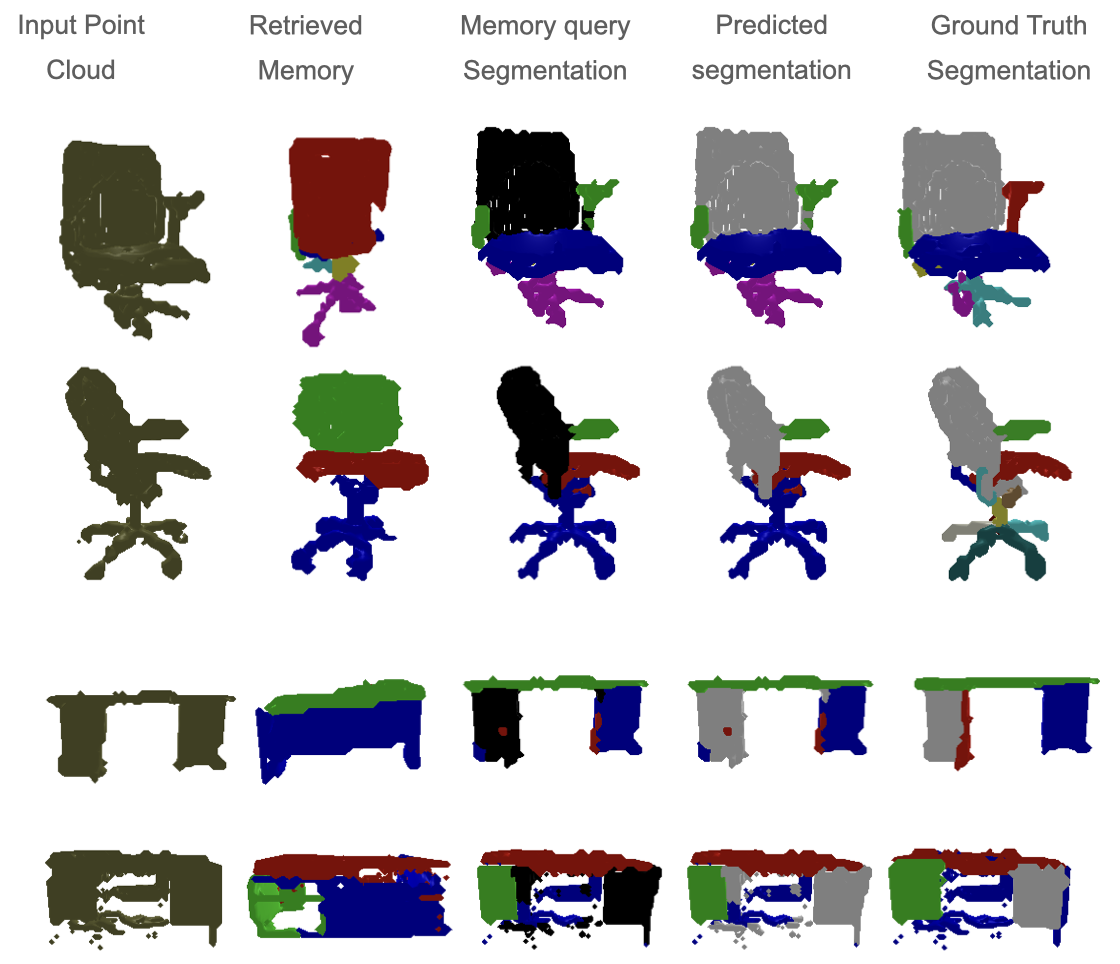}
  \caption{Results on base category samples from ScanObjectNN \cite{Uy2019RevisitingPC} using \model{}.} 
  \label{fig:viz_7}
\end{figure}

\begin{figure}[h!]
  \centering
    \includegraphics[width=0.8\linewidth]{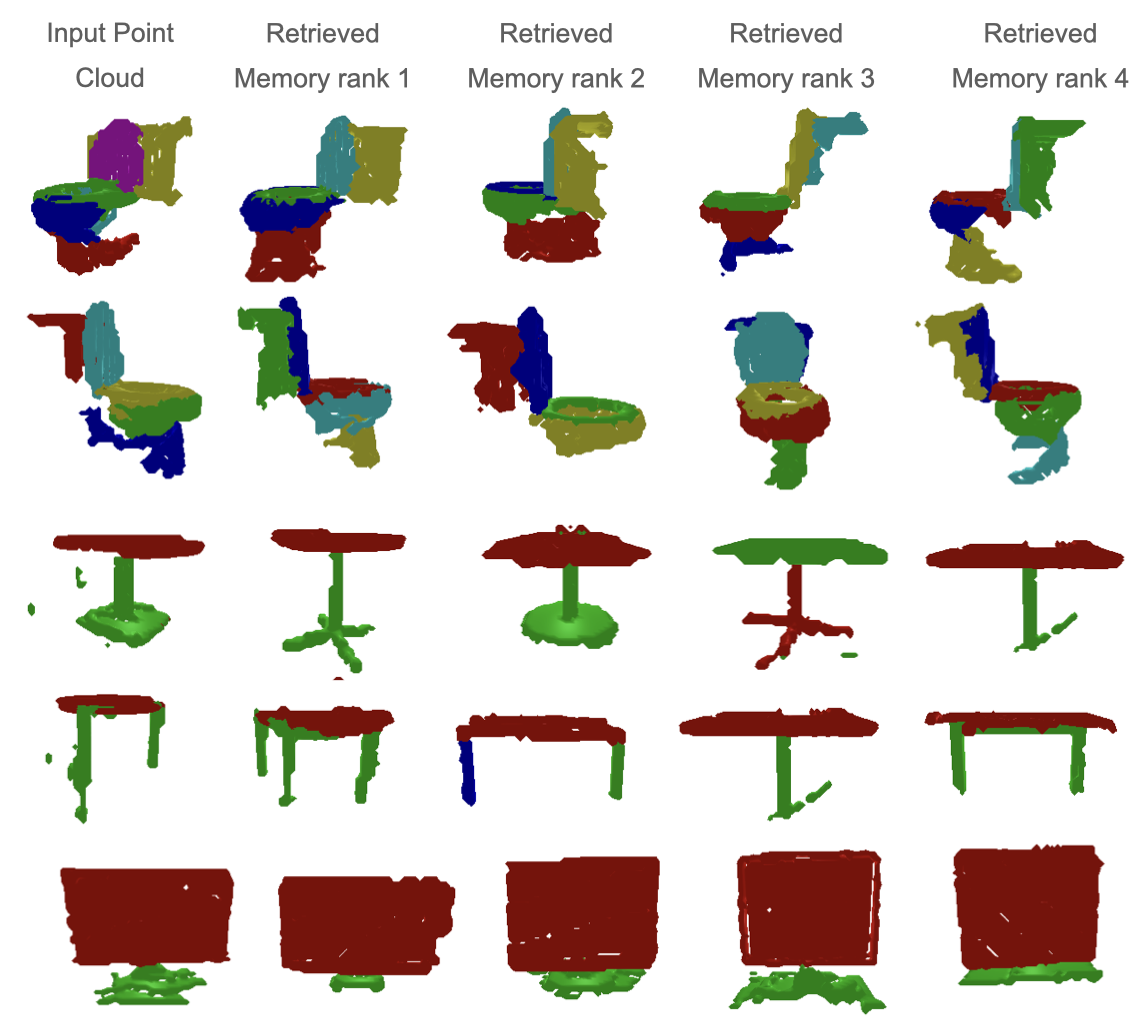}
  \caption{Top-$4$ retrieved results for the input point cloud from ScanObjectNN \cite{Uy2019RevisitingPC} dataset.}
  \label{fig:viz_retriever_3}
\end{figure}

We test \model{} on ScanObjectNN \cite{Uy2019RevisitingPC}, which contains noisy and incomplete real-world point clouds. We split the training into 11 classes seen during training (bag, bin, box, cabinet, chair, desk, door, pillow, shelf, sink, sofa) and 4 unseen (bed, display, table, toilet). Note that ScanObjectNN is not consistently labelled. For example, as we show in rows 5 and 6 of Figure \ref{fig:viz_7}, the legs of a chair may be annotated as a single part or multiple parts in the dataset. Although the PartNet dataset provides the ``level'' information, there is no such information in ScanObjectNN. Therefore we only qualitatively evaluate our model in Figure \ref{fig:viz_7} for base classes and in Figure \ref{fig:viz_8} for novel classes. We can see that our predictions are always plausible and consistent with the retrieved memory, even if the expected label space is different. We additionally visualize retrieval results in Figure \ref{fig:viz_retriever_3}.

\subsection{Qualitative Results for Single-Memory and Multi-Memory \model{}}
\label{parse_fig}

\begin{figure}[h!]
  \centering
  \includegraphics[width=1.0\linewidth]{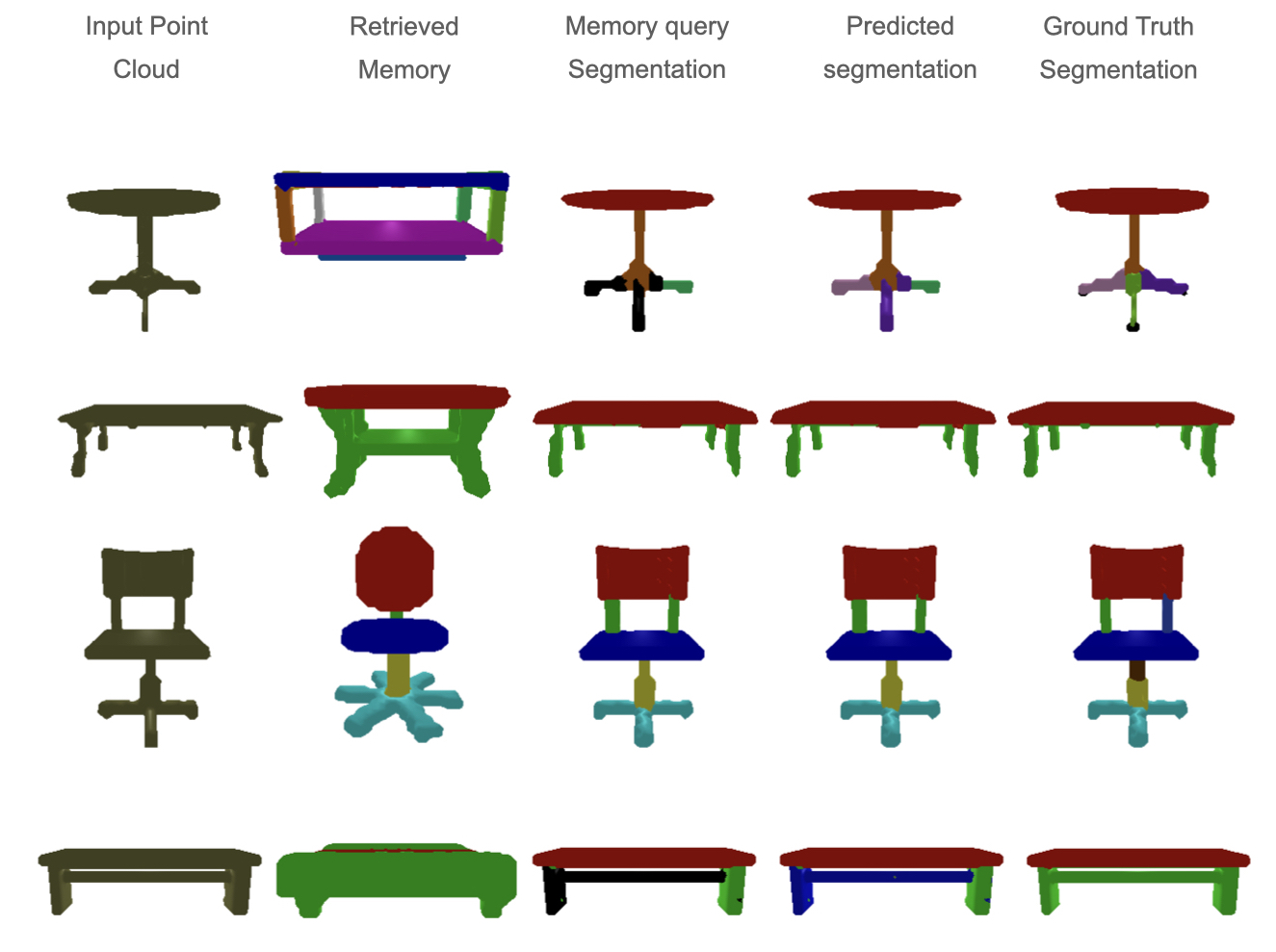}
    \includegraphics[width=1.0\linewidth]{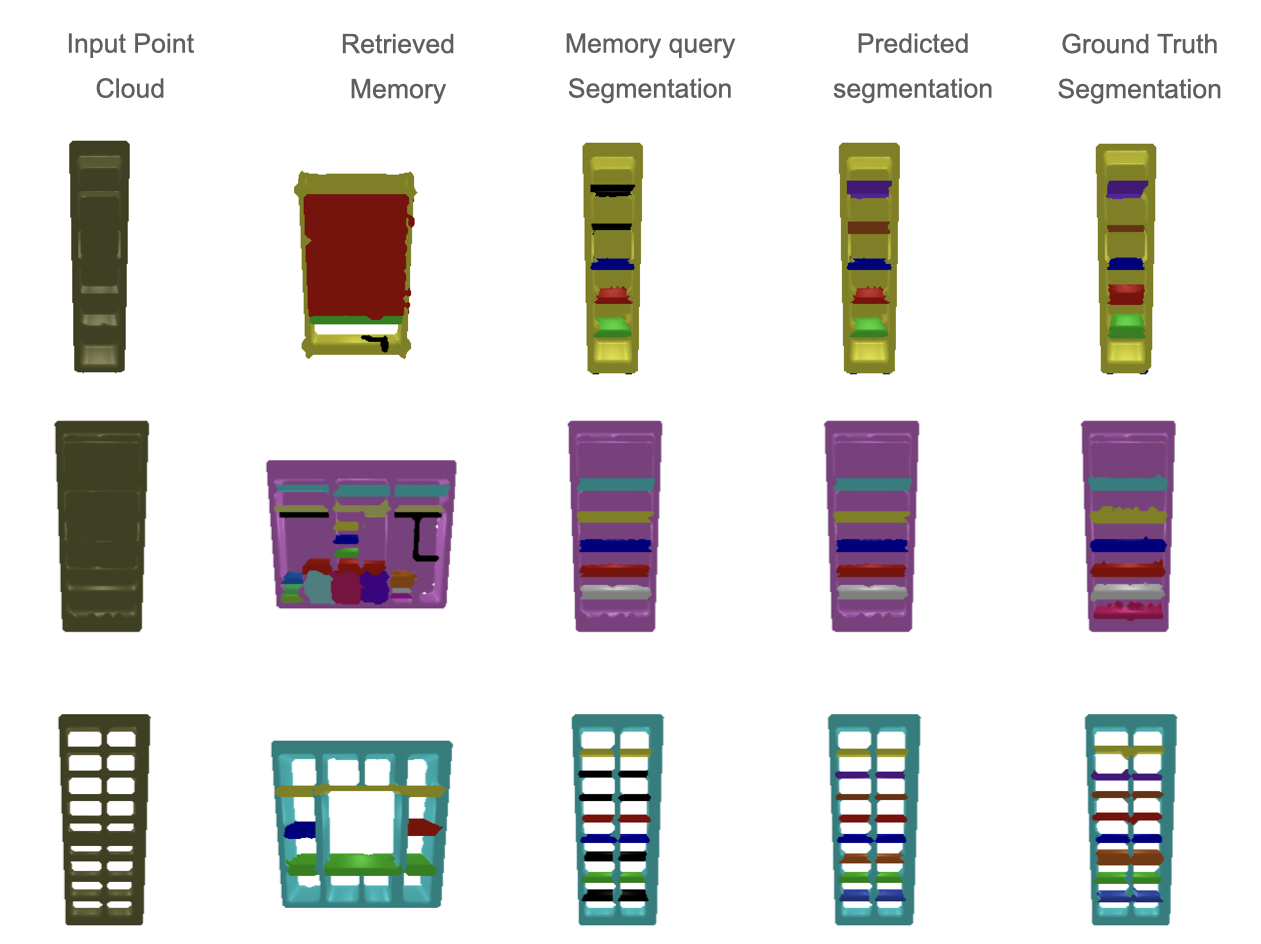}
  \caption{More qualitative object parsing results using \model{}.} 
  \label{fig:viz_1}
\end{figure}

\begin{figure}[h!]
  \centering
  \includegraphics[width=0.9\linewidth]{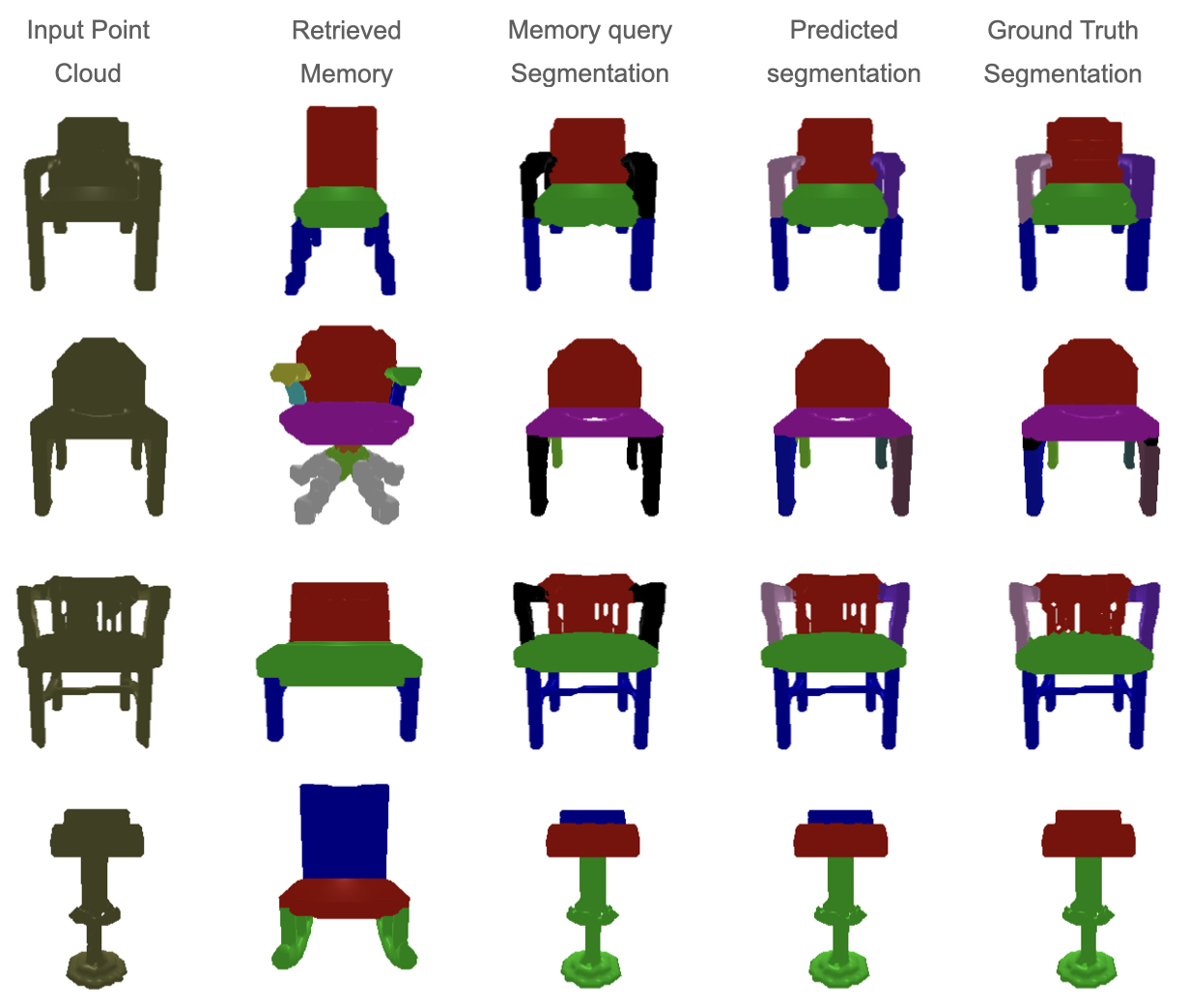}
\includegraphics[width=0.9\linewidth]{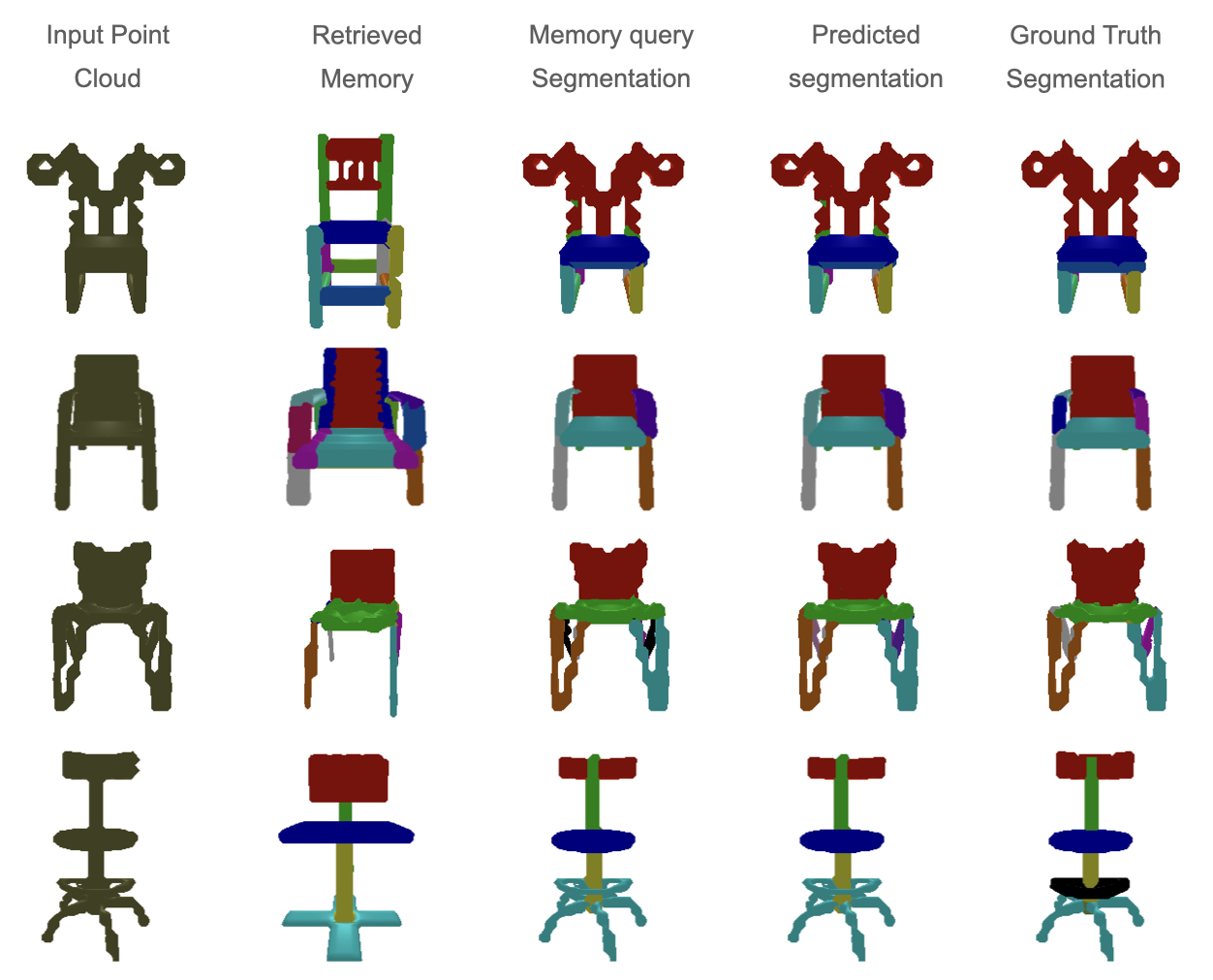}
  \caption{More qualitative object parsing results that are predicted by \model{}.} 
  \label{fig:viz_2}
\end{figure}

\begin{figure}[h!]
  \centering
  \includegraphics[width=0.9\linewidth]{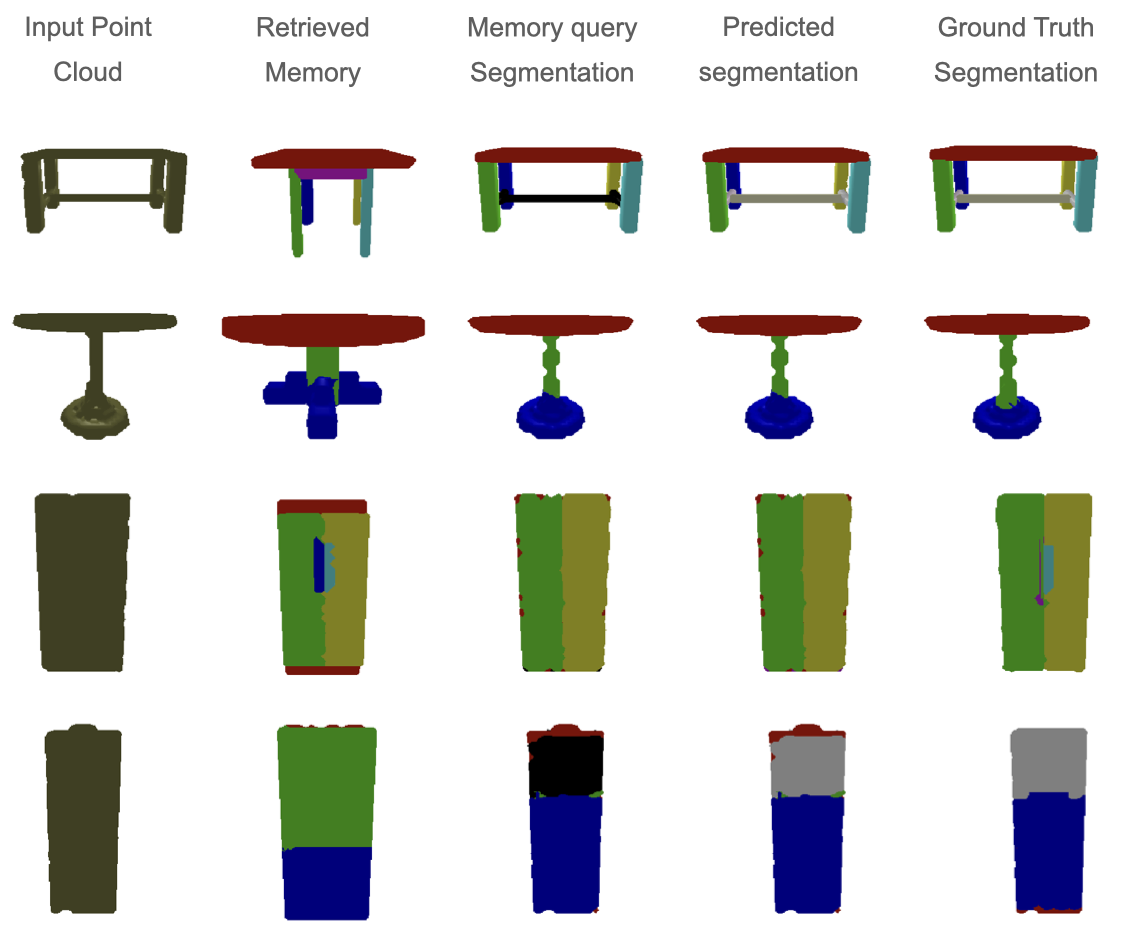}
    \includegraphics[width=1.0\linewidth]{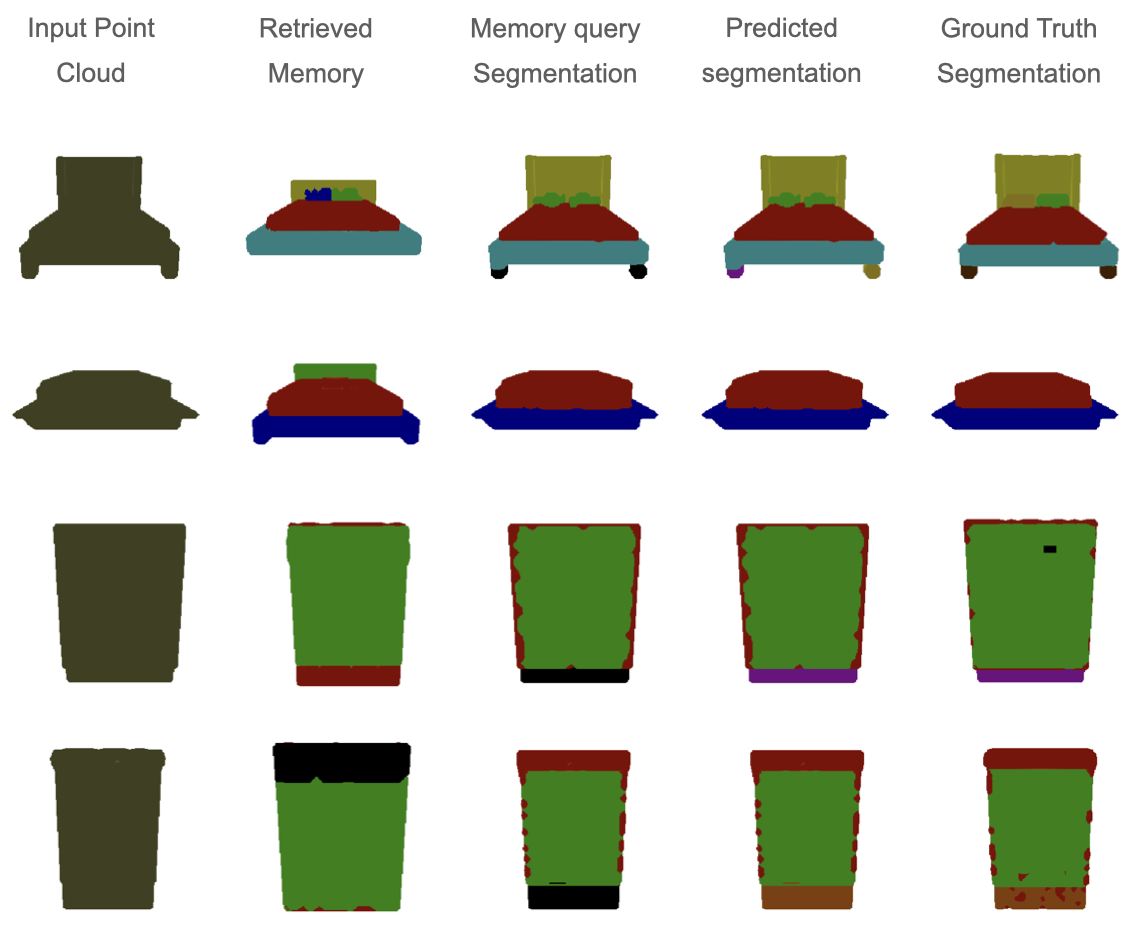}
  \caption{Qualitative results on novel category samples from PartNet dataset \cite{Mo_2019_CVPR} using \model{} \textbf{without fine-tuning}.} 
  \label{fig:viz_6}
\end{figure}

\begin{figure}[h!]
  \centering
  \includegraphics[width=1.0\linewidth]{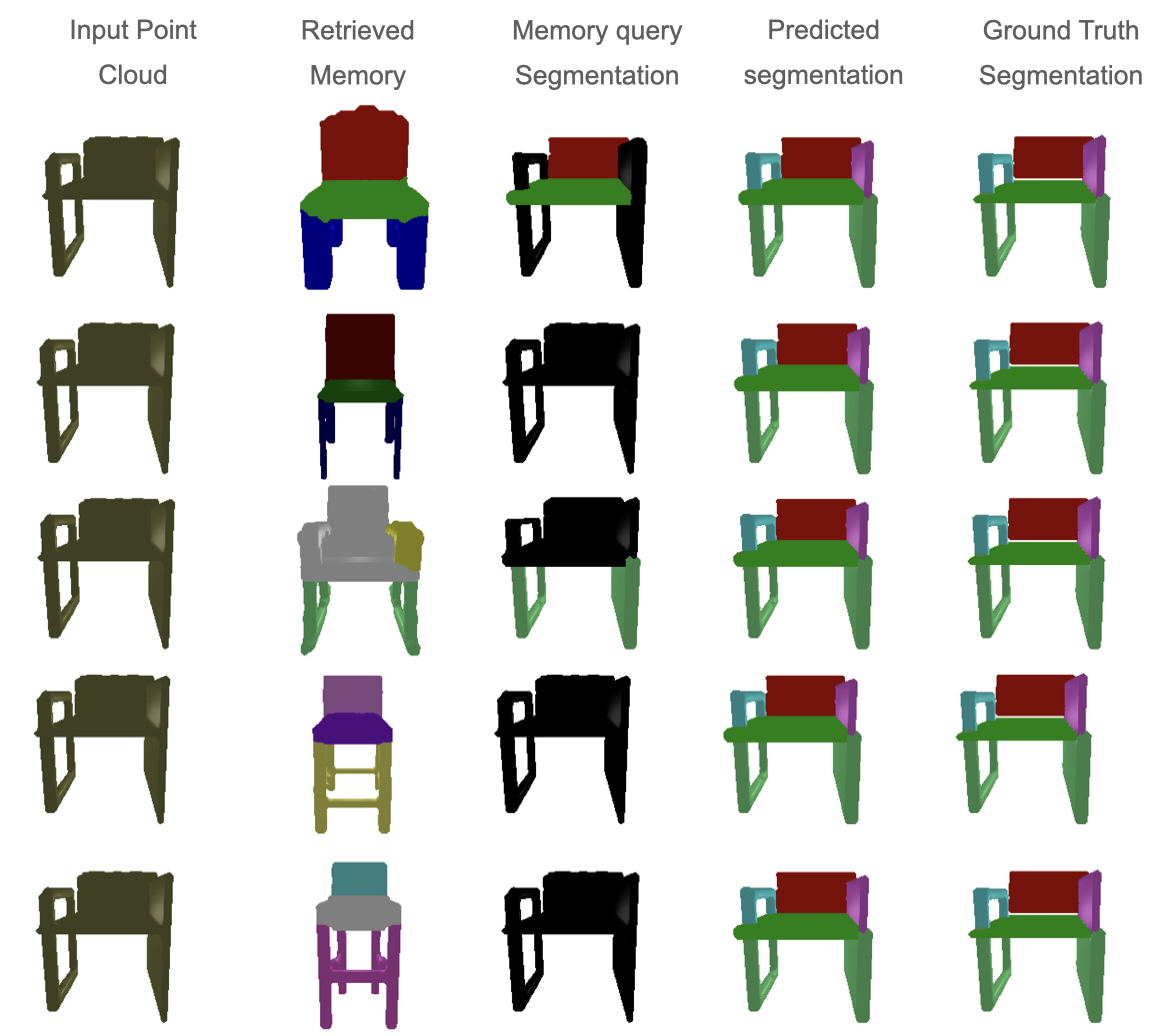}
  \caption{Modulation using multi-memory  \model{}. Our model takes as input $5$ different memories simultaneously and then parses the object. Each row shows the effect of a different memory. All memories decode simultaneously and we show which part each one decodes in the third column. In the fourth column we show the combined predictions of all memories and scene-agnostic queries.} 
  \label{fig:viz_3}
\end{figure}

\begin{figure}[h!]
  \centering
  \includegraphics[width=0.9\linewidth]{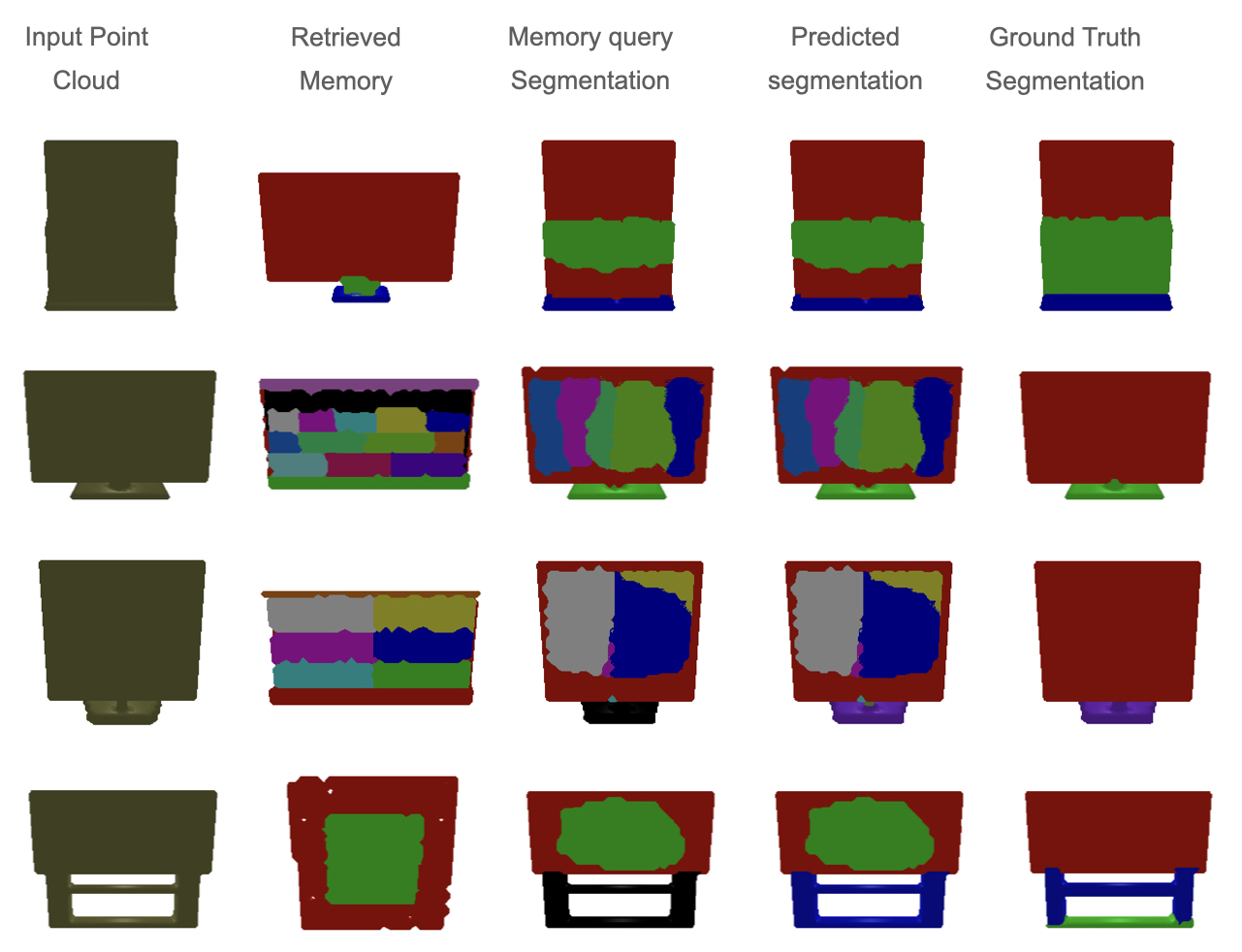}
    \includegraphics[width=1.0\linewidth]{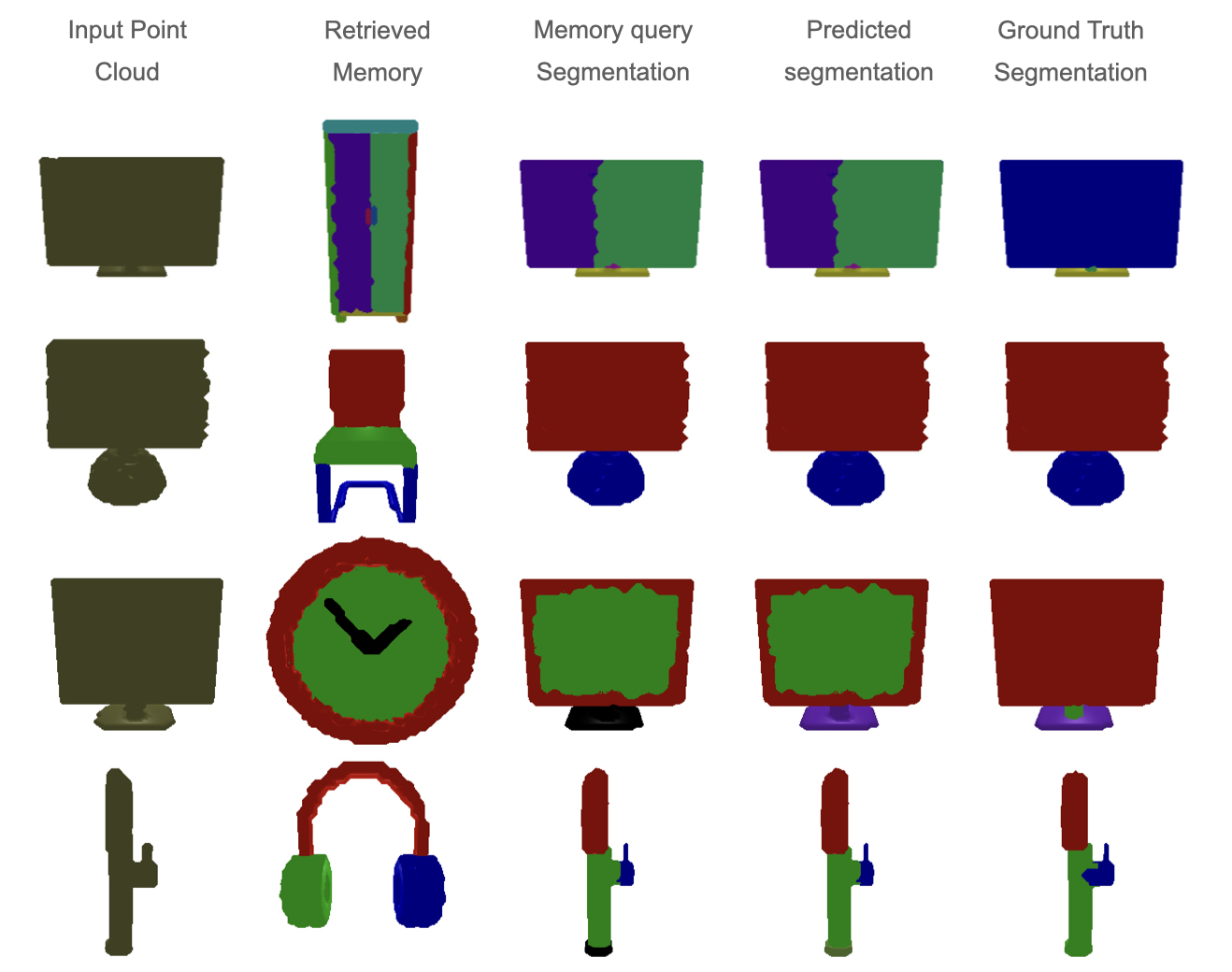}
  \caption{To qualitatively evaluate the effect of modulation in parsing: \textbf{Top}: we manually annotate the retrieved memory with random labels that do not correspond to any PartNet level. Our model adapts to the new label space. \textbf{Bottom}: we use as memory an object of a different category. The model is able to generalize geometric correspondences across instances of totally different classes, e.g. display and clock.} 
  \label{fig:viz_4}
\end{figure}

\begin{figure}[h!]
  \centering
  \includegraphics[width=1.0\linewidth]{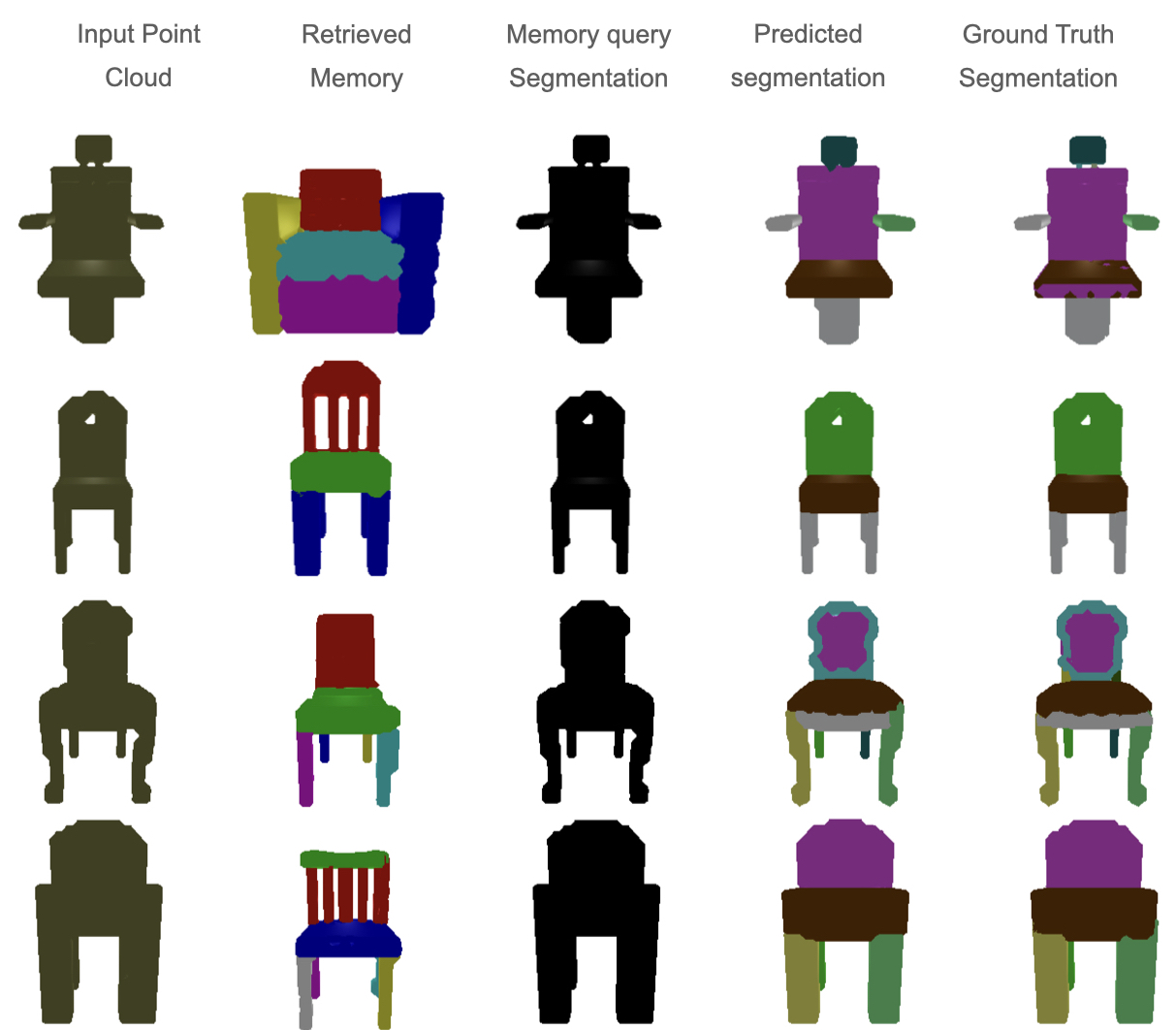}
  \caption{We show the parsing of input point cloud using \nopretrainsingle{}. Most regions are black in column 3, denoting that memory part queries do not decode anything and everything is being decoded by scene-agnostic queries. This highlights the role of within-scene pre-training for the emergence of part correspondence.} 
  \label{fig:viz_5}
\end{figure}

\begin{figure}[h!]
  \centering
    \includegraphics[width=0.8\linewidth]{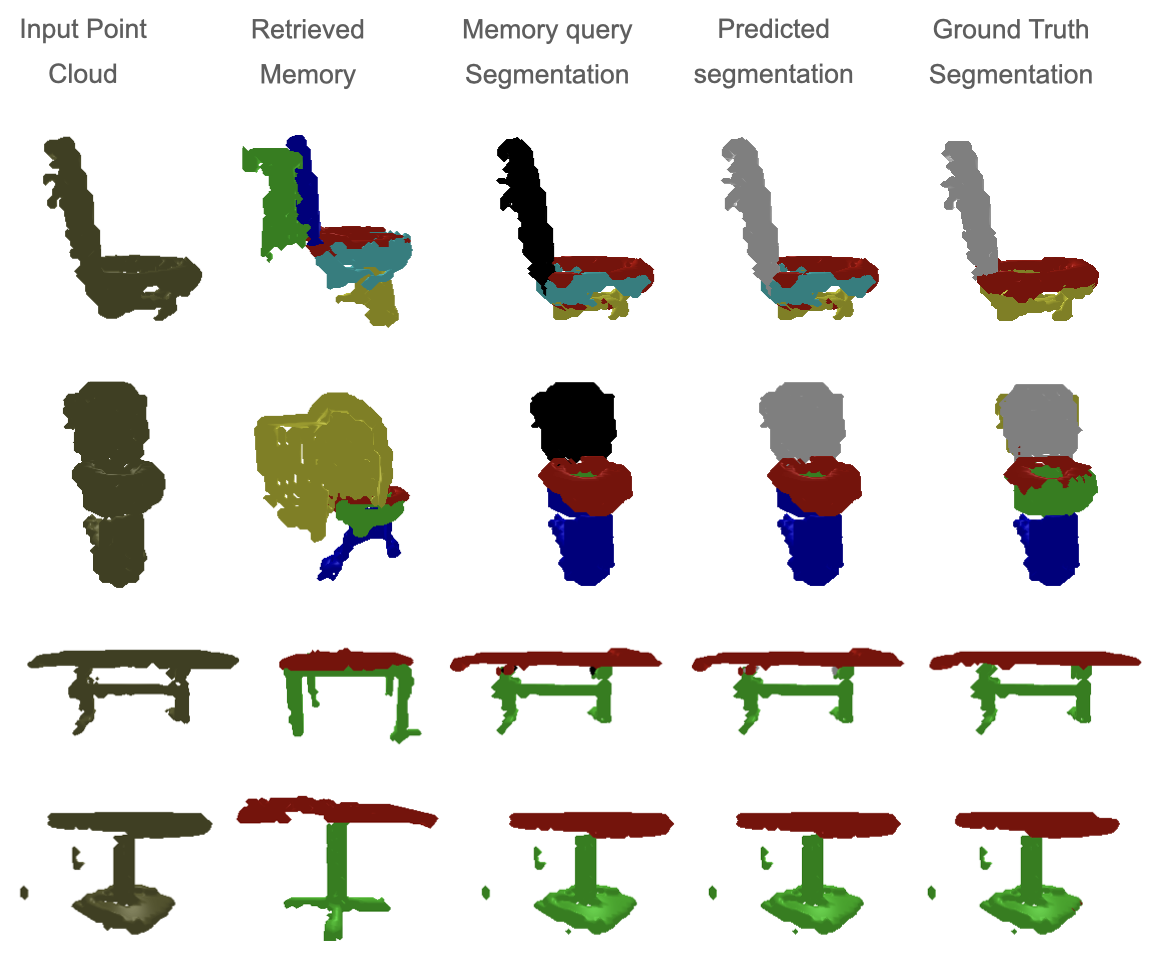}
  \caption{Results on novel category samples from ScanObjectNN \cite{Uy2019RevisitingPC} using \model{}.} 
  \label{fig:viz_8}
\end{figure}

In this section, we show our model's qualitative results for object parsing. In the Figures \ref{fig:viz_1}, \ref{fig:viz_2}, \ref{fig:viz_6}, \ref{fig:viz_3}, \ref{fig:viz_4}, \ref{fig:viz_5}, \ref{fig:viz_7}, \ref{fig:viz_8}  we use the following $5$-column pattern:
\begin{itemize}
    \item Unlabelled input point cloud
    \item Memory used for modulation
    \item Object parsing generated using only the memory part queries (not the scene-agnostic queries). In this column, regions that are not decoded by a memory part query are colored in black.
    \item Final predicted segmentation parsing using both memory-initialized queries and scene-agnostic queries. Regions that are colored in black in the third column but colored differently in the fourth column are decoded by scene-agnostic queries.
    \item Input point cloud's ground truth segmentation at the granularity level of the memory.
\end{itemize}

We qualitatively show the emergence of part correspondence between retrieved memory (column 2) and the input point cloud parsed using memory queries (column 3). Parts having the same color in columns 2 and 3 demonstrate correspondence, i.e. a part in column 2 decodes the part with the same color in column 3. \model{} promote correspondence of parts on both base (Figure \ref{fig:viz_1} bottom and \ref{fig:viz_2}) and novel (Figure \ref{fig:viz_1} top and \ref{fig:viz_6}) categories. This correspondence is semantic but also geometric, as can be seen in Figure \ref{fig:viz_4}. When multiple memories are available, \model{} mix and match parts of different memories to parse the input. Furthermore, we show parsing results for \nopretrainsingle{} in Figure \ref{fig:viz_5}. We observe that all of memory part query are inactive in the parsing stage. This demonstrates the utility of within-scene pre-training, as without this pre-training part correspondence does not emerge, as shown in Figure \ref{fig:viz_5}. Lastly, we show that \model{} generalize to noisy and incomplete point clouds in ScanObjectNN.

\subsection{Additional related work}\label{sec:additional_rel_work}

\paragraph{Neural-symbolic models}
\model{} are a type of neural-symbolic model that represents knowledge explicitly, in terms of structured visual memories,  where each one is a  graph of part-entity neural embeddings. 
A structured visual memory can be considered  the  neural equivalent of a FRAME introduced in \cite{minsky81framework},  \textit{``a graph of nodes and their relations  for representing a stereotyped situation, like being in a certain kind of living room, or going to a child's birthday party"}, to quote  \citet{minsky81framework}. 
FRAME nodes would operate as ``slots" to be filled  with specific entities, or symbols, in the visual scene. 
Symbol detection  would be carried out by a separate state estimation process 
such as general-purpose  object detectors \cite{https://doi.org/10.48550/arxiv.2201.02605}, employed also by recent neuro-symbolic models 
\cite{DBLP:journals/corr/abs-1810-02338, DBLP:journals/corr/abs-1904-12584, DBLP:journals/corr/abs-1910-01442}. 
In-the-wild detection of symbols (e.g., chair handles, faucet tips, fridge doors)  typically fails, which is  the  reason why these earlier  symbolic systems of knowledge, largely  disconnected from the sensory input, have not been widely adopted.  
\model{} take a step towards addressing these shortcomings by including symbol detection  as part of  inference itself,  through a top-down  modulation that uses the context represented in the  memory graph, to jointly search for  multiple entities and localize them in context of one another.

\textbf{3D instance segmentation} has been traditionally approached as a clustering problem \cite{Chen2009ABF,Sidi2011UnsupervisedCO}. Point-based methods learn either translation vectors mapping every point to its instance’s center \cite{Jiang2020PointGroupDP,Chen2021HierarchicalAF,Vu2022SoftGroupF3} or similarities across points \cite{Wang2018SGPNSG,Zhang2021PointCI}, followed by one or more stages of clustering. Similarly, \cite{Wang2021LearningFS,Jones2022SHRED3S} oversegment the point cloud into small regions and then merge them into parts. \citet{Yu2019PartNetAR} recursively decompose a point cloud into segments of finer resolution. 
\cite{Mo_2019_CVPR,Sun2022SemanticSI} learn representative vectors that form clusters by voting for each point. However, these approaches usually assume a fixed label space and need to train a separate model for each sub-task.
In contrast, we employ Detection Transformers \cite{Carion2020EndtoEndOD} for instance segmentation by repurposing the query vectors to act as representative vectors. We extend this set of queries with memory-initialized queries, enabling in-context reasoning. This allows us to train one model across all categories. As our results show, in absence of such memory contextualization, training one model across multiple categories hurts generalization (Table \ref{tab:results_single_cat}).

\end{document}